\documentclass[conference]{IEEEtran}
\IEEEoverridecommandlockouts
\usepackage[noadjust]{cite}
\usepackage{amsmath,amssymb,amsfonts}
\usepackage{algorithmic}
\usepackage{graphicx}
\usepackage{textcomp}
\usepackage{xcolor}
\def\BibTeX{{\rm B\kern-.05em{\sc i\kern-.025em b}\kern-.08em
    T\kern-.1667em\lower.7ex\hbox{E}\kern-.125emX}}

\usepackage{csquotes}
\ifCLASSOPTIONcompsoc
    \usepackage[caption=false, font=normalsize, labelfont=sf, textfont=sf]{subfig}
\else
    \usepackage[caption=false, font=footnotesize]{subfig}
\fi

\usepackage[roman]{parnotes}
\makeatletter
\def\parnoteclear{%
    \gdef\PN@text{}%
    \parnotereset
}

\usepackage{scalerel}
\usepackage{tikz}
\usetikzlibrary{svg.path}

\definecolor{orcidlogocol}{HTML}{A6CE39}
\tikzset{
  orcidlogo/.pic={
    \fill[orcidlogocol] svg{M256,128c0,70.7-57.3,128-128,128C57.3,256,0,198.7,0,128C0,57.3,57.3,0,128,0C198.7,0,256,57.3,256,128z};
    \fill[white] svg{M86.3,186.2H70.9V79.1h15.4v48.4V186.2z}
                 svg{M108.9,79.1h41.6c39.6,0,57,28.3,57,53.6c0,27.5-21.5,53.6-56.8,53.6h-41.8V79.1z M124.3,172.4h24.5c34.9,0,42.9-26.5,42.9-39.7c0-21.5-13.7-39.7-43.7-39.7h-23.7V172.4z}
                 svg{M88.7,56.8c0,5.5-4.5,10.1-10.1,10.1c-5.6,0-10.1-4.6-10.1-10.1c0-5.6,4.5-10.1,10.1-10.1C84.2,46.7,88.7,51.3,88.7,56.8z};
  }
}

\newcommand\orcidicon[1]{\href{https://orcid.org/#1}{\mbox{\scalerel*{
\begin{tikzpicture}[yscale=-1,transform shape]
\pic{orcidlogo};
\end{tikzpicture}
}{|}}}}

\usepackage{url}
\usepackage{hyperref}

\usepackage{flushend}

\hypersetup{
    colorlinks=true,
    allcolors= [rgb]{0,0.0,0.6},
    urlcolor=black
}

\usetikzlibrary{shapes.geometric, arrows,fit}
\usepackage{varwidth}

\tikzstyle{large_rect} = [rectangle, minimum width=2cm, minimum height=2.5cm, text centered, draw=black, very thick]
\tikzstyle{small_rect} = [rectangle, minimum width=2cm, minimum height=1.3cm, text centered, draw=black, very thick]

\tikzstyle{arrow} = [very thick,->,>=stealth]

\usepackage{siunitx}
\usepackage{afterpage}

\usepackage{makecell} %
\usepackage{booktabs}
\usepackage{array, tabularx, caption, boldline}
\usepackage{graphicx}
\usepackage{cellspace}
\usepackage[hashEnumerators,smartEllipses]{markdown}

\begin{document}

\title{A Review of Testing Object-Based Environment Perception for Safe Automated Driving*}

\author{\IEEEauthorblockN{
Michael Hoss$^{1} \orcidicon{0000-0001-9924-7596}$, 
Maike Scholtes$^{1} \orcidicon{0000-0003-2733-5292}$,
Lutz Eckstein$^{1}$
\thanks{*The research leading to these results is funded by the Federal Ministry for Economic Affairs and Energy within the project ``VVM - Verification and Validation Methods for Automated Vehicles Level 4 and 5". The authors would like to thank the consortium for the successful cooperation.\newline}
\thanks{$^{1}$The authors are with the research area Vehicle Intelligence \& Automated Driving, Institute for
Automotive Engineering, RWTH Aachen University, 52074 Aachen, Germany
{\tt\small \{hoss, scholtes, eckstein\}@ika.rwth-aachen.de}}%
}
}

\maketitle

 \makeatletter
 \renewcommand\@oddfoot{\hbox{}\rightmark\hfil\small{\thepage}}
 \makeatother

\begin{abstract}
Safety assurance of automated driving systems must consider uncertain environment perception.
This paper reviews literature addressing how perception testing is realized as part of safety assurance. 
We focus on testing for verification and validation purposes at the interface between perception and planning, and structure our analysis along the three axes 1) test criteria and metrics, 2) test scenarios, and 3) reference data. 
Furthermore, the analyzed literature includes related safety standards, safety-independent perception algorithm benchmarking, and sensor modeling. 
We find that %
the realization of safety-aware perception testing remains an open issue 
since challenges concerning the three testing axes and their interdependencies currently do not appear to be sufficiently solved. 

\end{abstract}

\section{Introduction}
\label{sec:intro}

To date, the safety assurance of automated vehicles of SAE Levels 4 and 5~\cite{SAE18} (AVs) has been an open issue, which is partially because the environment perception of AVs is subject to uncertainties.
AVs perceive their environment through a sensor setup that typically consists of at least
camera, radar, and lidar sensors. 
Perception algorithms, which typically use machine learning, %
then process and fuse the raw sensor data to a world model that is handed over to the subsequent functional modules. %
Typical world models contain a list of objects, where each object corresponds to a road user or other movable entity surrounding the AV~\cite{Dietmayer2014representation}.

Uncertainties in this world model can be caused by environmental influences, sensing hardware, perception software, and other factors.
These uncertainties can be divided into state, existence, and classification uncertainties~\cite{dietmayer2016predicting}. 
Respective examples are that a road user's velocity is estimated too slow, the road user is not detected at all, or a pedestrian is classified as a cyclist. 

\begin{figure}[t]
    \centering
    \vspace*{2mm}
    \includegraphics[width=\linewidth]{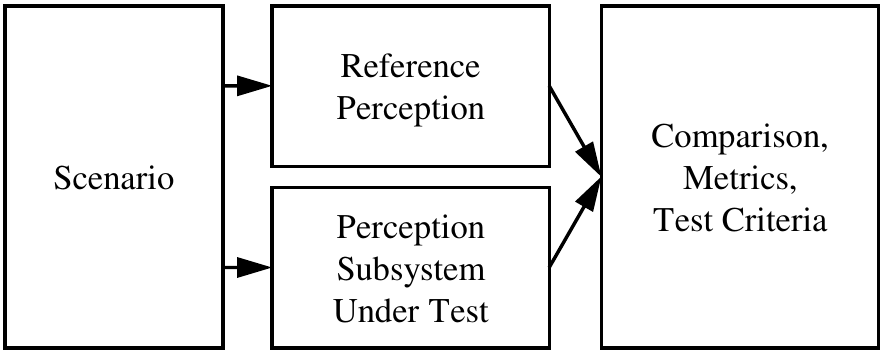}
    \caption{Taxonomy of Stellet et al.~\cite{stellet2015testing}, applied to perception testing: A real-world scenario is simultaneously observed by a reference perception system and by a perception subsystem under test. Their results are compared using metrics, which allows to evaluate test criteria.}
    \label{fig:taxonomy_general}
\end{figure}

For AV subsystems that are mostly defined by software, the real-world testing effort can be substantially reduced through simulated tests.  
However, as the sensor hardware interacts with the real environment in complex ways that are hard to replicate in simulations, real-world tests remain pivotal.

In the current literature, there does not yet exist a common methodology for the execution of real-world perception tests. 
Under which criteria is the environment perceived ``well enough"?
In which scenarios should the environment perception be tested?
How can one obtain the true environment state during these scenarios? 
Summarizing these questions, the primary research question of this review is: 

\begin{displayquote}
\textit{How can perception tests be realized to assess that an uncertain perception subsystem allows safe AV behavior?}
\end{displayquote}

To structure the diverse literature contributions to this general question, we apply the axes of the testing taxonomy by Stellet et al.~\cite{stellet2015testing} to perception testing %
(Fig.~\ref{fig:taxonomy_general}). 
The taxonomy states that testing is an evaluation of

\begin{displayquote}
``A statement on the system-under-test (\textit{test criteria}) that is expressed quantitatively (\textit{metric}) under a set of specified conditions (\textit{test scenario}) with the use of knowledge of an ideal result (\textit{reference})."
\end{displayquote}

\begin{figure*}[t]
    \centering
    \vspace*{2mm}
    \includegraphics[width=\linewidth]{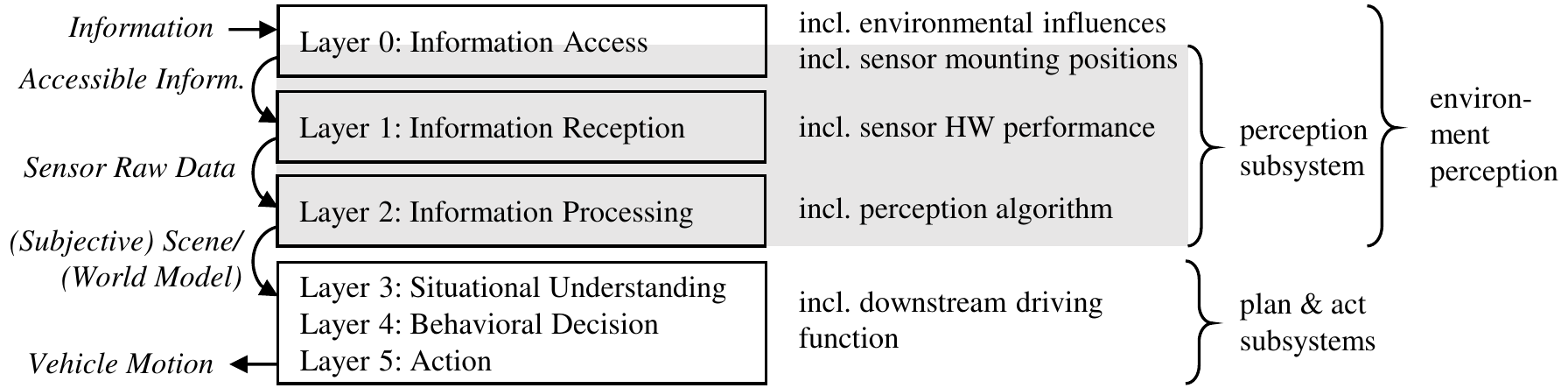}
    \caption{Left: %
    Functional decomposition layers and interfaces of Amersbach and Winner~\cite{amersbach2017functional}, focused on perception. 
    Right: terms that are used in this paper, mapped to the layers of~\cite{amersbach2017functional}. %
    }
    \label{fig:amersbach_layers}
\end{figure*}

\subsection{Motivation for a Review}

The current literature from the areas of perception development, automotive safety assurance, and safety in artificial intelligence contains valuable approaches to all mentioned testing axes. 
For a successful safety proof, approaches to individual axes have to seamlessly fit into an overall methodology. 
Prior to writing, we found it difficult to identify how the available approaches complement each other, 
which is why we are aiming at providing a structured overview of the state of the art. 
The goal of this review is to enable the development of holistic concepts for testing the safety-relevant uncertainties of environment perception. %

\subsection{Structure and Contributions}
Related Reviews, Surveys, and Overviews (Sec.~\ref{sec:related_work}) reviews previous secondary literature. 
Methods (Sec.~\ref{sec:methods}) defines the scope and the literature search process. 
Our main contributions are overviews on:

\begin{itemize}
    \item Perception Testing in Safety Standards (Sec.~\ref{sec:standards}) %
    \item Established perception testing that does not directly target the primary research question (Sec.~\ref{sec:established})
    \begin{itemize}
    \item Perception Algorithm Benchmarking (Sec.~\ref{sec:developer_testing})
    \item Object-Lvl. Data-Driven Sensor Modeling (Sec.~\ref{sec:sensor_modeling})
    \end{itemize}
    \item Approaches to the research question, structured by the three individual testing axes
    \begin{itemize}
    \item Test Criteria and Metrics (Sec.~\ref{sec:axis_criteria_metrics})
    \item Test Scenarios (Sec.~\ref{sec:axis_test_scenarios})
    \item Reference Data (Sec.~\ref{sec:axis_ref_data})
    \end{itemize}
\end{itemize}

Finally, the Discussion of Review Results (Sec.~\ref{sec:discussion}) %
highlights the largest issues for answering the research question,
points out intersection topics between the testing axes, 
and puts the topics of this review in a larger context. %

\begin{table*}[t]
\centering
\caption{Thematic focus of this review paper}
\label{table:focus_of_paper}
\begin{tabularx}{\linewidth}{
>{\hsize=0.35\hsize}X 
>{\hsize=0.75\hsize}X 
>{\hsize=0.85\hsize}X 
} 
 \toprule
 \parnoteclear %
\textbf{Category} & \textbf{Focused aspect} & \textbf{Related aspects outside of main focus} \\
\midrule
Location in V-model %
& Right-hand side (offline testing for verification and validation purposes)
& Left-hand side; developer testing; runtime monitoring; data gathering for requirements identification~(\cite{Koopman2018towardFramework}); shadow mode \\
\hline
 Test method and target 
 & Black-box testing of interface between perception and planning; description of perception output
 &  White-box testing inside the perception or planning subsystems; %
 description of planning sensitivity  \\ 
\hline
Safety concept 
 & Safety of the intended functionality (SOTIF, see~\cite{iso2019sotif}), specifically object and event detection and response (OEDR) for collision mitigation
 &  Functional safety according to ISO 26262;
 Extension of ISO~26262 w.r.t. machine learning; cybersecurity; physical harm caused by sensors \\
 \hline 
Open/closed loop testing
 & Open-loop testing (enables offline testing of various perception algorithms on recorded raw sensor data) %
 & Closed-loop testing
 \\
 \hline
 Environmental entity 
 & Moving and non-moving other road users
 & Ego vehicle localization, static obstacles, road infrastructure, traffic signs, traffic lights, drivable space, auditory information \\
 \hline 
 Environment representation format 
 & Object list
 & Sensor raw data, object-independent environmental features, occupancy grid map, parametric drivable space \\
\bottomrule
\end{tabularx}
\parnotes
\end{table*}

\subsection{Term Definitions}
\label{sec:definitions}

The following term definitions are used throughout the rest of this paper. %
Fig.~\ref{fig:amersbach_layers} illustrates the relations between different terms by using the the layered decomposition of an automated driving (AD) system by~\cite{amersbach2017functional}. 

\subsubsection{Perception Subsystem}
\label{sec:def_perception_subsystem}
The AV's subsystem for environment perception is defined by at least sensor mounting positions (part of Layer~0), the sensors themselves that generate raw data (Layer~1), and perception software (Layer~2) (Fig.~\ref{fig:amersbach_layers}).

\subsubsection{System Under Test}
\label{sec:def_sut}
Perception subsystem of a subject vehicle/ego vehicle. Abbreviation: SUT.

\subsubsection{Environment Perception}
\label{sec:def_environment_perception}
The process of environment perception consists of the mentioned perception subsystem, which is being affected by environmental influences that are outside the control of it (rest of Layer~0; Fig.~\ref{fig:amersbach_layers}). 
These environmental influences determine if and how the individual parts of the ground truth information are accessible~\cite{amersbach2017functional}. 

\subsubsection{Perception Algorithm}
\label{sec:def_perception_algo}
Software that computes an object-based world model from the raw sensor data (adapted from~\cite{VVM2020Begriffsregister}). 
Generally includes both traditional code and components that are trained on data. 
Corresponds to Layer~2 (Fig.~\ref{fig:amersbach_layers}).

\subsubsection{World Model}
\label{sec:def_world_model}
Representation of the ego vehicle's environment and own state that is computed by a perception algorithm (adapted from~\cite{VVM2020Begriffsregister}).
Equal to a subjective scene at the interface between Layers 2 and 3 (Fig.~\ref{fig:amersbach_layers}). 
The most essential world model component in this paper is the current state of dynamic object tracks.

\subsubsection{Safety}
``absence of accidents, where an accident is an event involving an unplanned and unacceptable loss"~\cite[p.\,11]{leveson2016engineering}.  %
Unless stated otherwise, safety in this paper means \textit{safety of the intended functionality} (SOTIF) of the AV (see~\cite{iso2019sotif}) rather than functional safety according to ISO 26262~\cite{ISO_26262_2018}.

\subsubsection{Reliability}
\label{sec:def_reliability}
``probability that something satisfies its specified behavioral requirements over time and under given conditions -- this is, it does not fail"~\cite[p.\,10]{leveson2016engineering}.  %

\subsubsection{Uncertainty}
\label{sec:def_uncertainty}
``doubt about the validity of the result of a measurement", or a quantitative measure of this doubt~\cite{JCGM2008GUM}. 
Uncertainty in object-based world models can be state, existence, or classification uncertainty~\cite{dietmayer2016predicting}, as mentioned before.

\subsubsection{Scenario}
\label{sec:def_scenario}
``A scenario describes the temporal development between several scenes in a sequence of scenes. [...]" \cite{Ulbrich2015}. 
For open-loop perception testing, we assume that the temporal development of all ground truth scenes is given, including environmental influences and all road user trajectories.
What is to be determined in a test scenario is how the SUT subjectively observes the objective scenes in its world model (more in Sec.~\ref{sec:what_are_perc_scenarios}).

\subsubsection{Metrics}
\label{sec:def_metric}
Functions that compute properties of interest about the SUT.
We use the microscopic/macroscopic terminology for metrics of~\cite{Junietz2019DissRiskMetrics}, where \textit{microscopic} refers to a single scene for a single ego vehicle within a single scenario, and \textit{macroscopic} refers to the average over an entire fleet of identical vehicles that encounters various scenarios and scenes.

\section{Related Reviews, Surveys, and Overviews}
\label{sec:related_work}

Previous reviews partially cover the targeted topic of this paper.
For example, the surveys by Stellet et al.~\cite{Stellet2020ValidationSurvey},~\cite{stellet2015testing}, Riedmaier et al.~\cite{Riedmaier2020_surveySBT}, Nalic et al.~\cite{Nalic2020sbt_survey}, and the PEGASUS method~\cite{Pegasus2019Overview} deal with testing and safety assurance of AVs in general, but without specifically analyzing the interface between perception and planning. 

A systematic literature review of the Swedish SMILE project targets verification and validation for machine learning based systems in AD~\cite{borg2018safely}, however, without focusing specifically on environment perception. 

Literature reviews about AV sensing and perception like~\cite{marti2019review},~\cite{Rosique2019}, and~\cite{Zhu2017PerceptionOverview} provide the technological state of the art, but do not explicitly focus on testing or safety assurance. 

Furthermore, there are publications that provide a valuable overview over current safety challenges for AV environment perception, but which are not explicitly literature reviews~\cite{Willers2020safety},~\cite{Burton2017SafetyML_HAD}.
For example,~\cite{Willers2020safety} identifies safety concerns and mitigation approaches for safety-critical perception tasks that rely on deep learning. %

In summary, we aim to fill the existing gap in literature reviews by analyzing literature related to the main research question in depth throughout the rest of this paper.

\section{Methods}
\label{sec:methods}

The methods of this literature review are loosely inspired by the guidelines for systematic literature reviews~\cite{Kitchenham2007guidelinesSLR} and snowballing search~\cite{wohlin2014snowballing}, and respective example applications of both in~\cite{Schalling2019LidarBenchmarking} %
and~\cite{borg2018safely}.
The suggested guidelines are purposely not followed exactly, because they were designed for reviewing quantitative and empirical research rather than qualitative or position papers, which are, however, common in the current automotive safety assurance domain.

\begin{table*}[t]
\centering
\caption{Iterations of snowballing literature search. *Considered leaf search results without further search because of thematic distance to primary research question.}
\label{table:snowballing_iterations}
\begin{tabularx}{\linewidth}{
>{\hsize=0.15\hsize}X 
>{\hsize=1.85\hsize}X 
} 
 \toprule
 \textbf{Snowballing iteration} & 
 \textbf{Found sources for further forward and backward snowballing search (58 publications total)} \\
 \midrule
 Start set 
 \newline 
 & {
 \textbf{From undocumented search:}
 Salay et al.~\cite{Salay2020_purss},~\cite{Salay2019_safety_perceptual_components},
 Salay and Czarnecki~\cite{Salay2019partialspecifications},
 Czarnecki and Salay~\cite{Czarnecki_2018_Framework},
 Berk et al.~\cite{berk2019exploiting},~\cite{Berk2019ReferenceTruth}, %
 Shalev-Shwartz et al.~\cite{shalevshwartz2017formalRSS},
 Amersbach and Winnner~\cite{amersbach2017functional},
 Stellet et al.~\cite{Stellet2020ValidationSurvey}, 
 Brahmi et al.~\cite{brahmi2013reference}, 
 Cassel et al.~\cite{Cassel2020SAEPerceptionRequirements};
 \newline 
\textbf{From undocumented and keyword search:} 
 Berk~\cite{Berk2019Dissertation}, 
 Stellet et al.~\cite{stellet2015testing},
 Salay et al.~\cite{Salay2017MLIso26262},
 Ulbrich et al.~\cite{Ulbrich2015};
 \newline
\textbf{From keyword search:} 
Borg et al.~\cite{borg2018safely},
Willers et al.~\cite{Willers2020safety}, 
Burton et al.~\cite{Burton2017SafetyML_HAD}, 
Ta{\c{s}} et al.~\cite{tas2016functional}, 
Kaprocki et al.~\cite{kaprocki2019multiunit}, 
Martin et al.~\cite{Martin2019}, 
Bai et al.~\cite{Bai2019_external_influence_factors_sensing} (22 publications).
} 
\\ 
\hline
Iteration 1 & 
Ulbrich et al.~\cite{ulbrich2017functional}*,
Althoff~\cite{althoff2010reachability}*,
Bagschik et al.~\cite{Bagschik2018_ontology}*,
Burton et al.~\cite{Burton2019confidence_arguments}, 
Koopman and Fratrik~\cite{Koopman2019_howMany}, 
\cite{Koopman2018towardFramework},
Kubertschak et al.~\cite{Kubertschak2014kamerabasiertes},
Strigel et al.~\cite{Strigel2014_KoPerDataset},
Berk et al.~\cite{Berk2020summary}, %
\cite{Berk2017BayesianTestDesign}*,
\cite{Berk2019rainfall}*,
Johansson and Nilsson~\cite{Johansson2016perceptionASIL}*,
Johansson et al.~\cite{Johansson2017assessingUseOfSensors}*,
Bock et al.~\cite{Bock2016testEffortEstimation}*,
Rivero et al.~\cite{Rivero2017roadDirtLidar}*,
Rasshofer et al.~\cite{Rasshofer2011influences}*, 
Fawcett~\cite{Fawcett2006ROC}*,
Cao and Huang~\cite{cao2018application},
Czarnecki et al.~\cite{Czarnecki2018OntologyPart2}*,
Meyer~\cite{Meyer1992designByContract}*,
Guo~\cite{Guo2017calibration}*,
Salay and Czarnecki~\cite{Salay2018UsingMLSafely}, %
Schwaiger et al.~\cite{Schwaiger2020uncertainty},
Arnez et al.~\cite{Arnez2020_uq_comparison},
Piazzoni et al.~\cite{Piazzoni2020Modeling_conference},
Weast~\cite{Weast2020},
Klamann et al.~\cite{Klamann2019}*,
Aravantinos and Schlicht~\cite{Aravantinos2020}*,
Breitenstein et al.~\cite{Breitenstein2020visual}*
(29 publications).
\\
\hline
 Iteration 2 %
 &  
Henne et al.~\cite{henne2020benchmarking},
Fleck et al.~\cite{Fleck2019testAreaBW},
Spanfelner et al.~\cite{Spanfelner2012ChallengesISO26262}*,
Rahimi et al.~\cite{Rahimi2019requirements},
Czarnecki~\cite{Czarnecki2019devOps}*,
Feng et al.~\cite{Feng2018uncertainty}*,~\cite{Feng2019calibrationECE} (7 publications).
\\ 
\bottomrule
\end{tabularx}
\end{table*}

\subsection{Thematic Scope}

Table~\ref{table:focus_of_paper} summarizes the focus of this literature review. 
Literature that matches more focused aspects is more likely to be included and discussed in detail, whereas literature about related aspects outside the main focus might still be referenced to provide context where the actually targeted literature was sparse. 

\subsection{Literature Search Process}

Three search processes were performed to accumulate the references of this paper. 
First, undocumented searches provided an initial knowledge base. 
Second, a keyword-based search complemented the starting set for a final iterative snowballing search. 
While the undocumented search provided most references about standards and established testing approaches, the inclusion criteria of the documented search targeted specifically the primary research question and its test axes. 

\subsubsection{Undocumented Search}

All references that are not part of Table~\ref{table:snowballing_iterations} were obtained through undocumented search processes before and during the writing process. %

\subsubsection{Keyword Search}

The following strings were searched in Google Scholar, allowing results in the time frame between 01/01/2010 and 20/08/2020:

\begin{itemize}
    \item (``automated driving" OR automotive) AND (perception OR sensing) AND (verification OR validation OR safety OR reliability OR sotif OR standard OR requirements OR specification OR testing OR metric)
    \item (``automated driving" OR automotive) AND (perception OR sensing) AND (reference OR data OR test OR criteria)
\end{itemize}
For both search strings, the top 250 results were analyzed.
20 publications were identified as candidates for the start set of the subsequent snowballing search, out of which 11 were actually included (see Table~\ref{table:snowballing_iterations}).

\subsubsection{Snowballing Search}

The results from the other search processes formed a start set, based on which we performed a forward and backward snowballing search~\cite{wohlin2014snowballing} in fall 2020 using Google Scholar to fill possible gaps. 
Forward snowballing means searching through the citations that a publication has received, while backward snowballing means searching through the references it itself has included. 
Both directions were searched within the same iteration, respectively (Table~\ref{table:snowballing_iterations}).

Included snowballing search results that are not directly related to the primary research question or to at least one of the three testing axes were excluded from the further search (marked with * in Table~\ref{table:snowballing_iterations}) to keep the search process efficient and the overall effort feasible.
Eventually, we removed the third iteration, which consisted of two sources, and some sources from iterations 1 and 2 due to irrelevance for the final version. 

\section{Perception Testing in Safety Standards}
\label{sec:standards}
Safety standards, guidelines, and frameworks can prescribe certain aspects around perception testing in safety assurance. 
Thus, we first review the explicit role of perception testing in standardizing literature before moving on to more concrete realizations of perception tests in the subsequent sections.

\subsection{ISO 26262} %

ISO 26262~\cite{ISO_26262_2018} addresses the assurance of functional safety of electric and electronic systems in the automotive domain. 
It has been the most essential safety standard for advanced driver assistance systems (ADAS), but has not been developed to assure the safety of AD systems that rely on environment perception or use machine learning~\cite{Salay2017MLIso26262}.
Consequently, the literature investigates how ISO 26262 could be adapted or extended to also cover machine-learning-based components such as modern environment perception systems, see e.g.~\cite{borg2018safely}. 

One major issue that prevents the concepts of ISO 26262 from being applied to AV environment perception is that the perception subsystem is said to be not fully specifiable (except for defining a training set), while ISO 26262 implicitly assumes that all functionality must be specified~\cite{Salay2017MLIso26262},~\cite{Salay2018UsingMLSafely}. 
This means that the environment perception is also not fully verifiable and perception testing is in general incomplete~\cite{Salay2017MLIso26262}.

Nevertheless, ISO 26262's workflow for Safety Elements out of Context (SEooC)~\cite[Part~10]{ISO_26262_2018} might be a hint on how the development of a perception subsystem (safety element) could at least be generally organized while the rest of the driving function (context) is not known.

\subsection{ISO/PAS 21448} %
A vehicle that is perfectly functionally safe according to ISO 26262 can still cause accidents if its behavioral logic is wrong~\cite{shalevshwartz2017formalRSS}. 
Addressing this issue, ISO/PAS 21448~\cite{iso2019sotif} introduces the concept of the so-called Safety of the Intended Functionality (SOTIF). 
The standard describes how SOTIF should be assured for ADAS up to SAE Level 2 and acknowledges that it is likely not sufficient for higher levels of automation. 

According to~\cite[Clause~7]{iso2019sotif}, so-called \textit{triggering events} ``that can trigger potentially hazardous behavior shall be identified", and their impact on the SOTIF shall be assessed. 
Triggering events related to sensor disturbances can be caused for example by poor weather conditions or poor-quality reflections. 
They can cause errors in the world model if corresponding \textit{functional insufficiencies} in the perception subsystem exist~\cite{Willers2020safety}.
Once triggering events for the perception subsystem have been identified (see e.g.~\cite{Martin2019}, \cite[Annex~F]{iso2019sotif}) as part of the so-called sensor verification strategy~\cite[Sub-Clause~10.2]{iso2019sotif}, they can also serve as test scenarios for it. 
Furthermore, \cite[Annex~D]{iso2019sotif} lists non-comprehensive example aspects to consider in verification testing of the perception subsystem, but leaves open the concrete realization of those tests. 

\subsection{NHTSA Vision and Framework}

In ``Automated Driving Systems 2.0: A Vision for Safety"~\cite{NHTSA2017vision_for_safety2}, the National Highway Traffic Safety Administration (NHTSA) and the U.S. Department of Transportation (DOT) follow a ``nonregulatory approach to automated vehicle technology safety".
AV manufacturers are allowed and %
required to follow and document their own individual safety assurance processes, which intends to support innovation rather than hindering it. 
Twelve so-called safety design elements are mentioned, where the one most relevant to this paper is \textit{Object and Event Detection and Response} (OEDR) -- the capability to ``detect any circumstance that is relevant to the immediate driving task" and respond to it.

A more recent and more detailed publication of NHTSA defines a framework for testing AD systems~\cite{Thorn2018NHTSA}. 
This framework also addresses OEDR and further defines a taxonomy that can help describing an operational design domain (ODD) for an AV.
Such a clear ODD description is necessary to express the requirements of how to also cope with unusual situations such as emergency vehicles. 

According to~\cite[p.\,65]{Thorn2018NHTSA}, testing the perception outputs can significantly facilitate the assessment of OEDR capabilities as compared to testing only the resulting trajectories. 
It can answer questions such as at which range obstacles are detected, if obstacles are correctly classified, and if their location and size are correctly estimated.
However, this NHTSA document also leaves open many details of technical realization of such tests.

\subsection{UL4600}
Another approach to achieve a safety standard is UL4600 by Underwriters Laboratories and Edge Case Research~\cite{UL4600_voting_2019} (voting version). Its full name is \textit{Standard for Safety for the Evaluation of Autonomous Products}. UL4600 is a voluntary industry standard for autonomous products and thus also applies to AD functions.
The creation of this standard is described in~\cite{Koopman2019}. 

Similar to the previously addressed standards, UL4600 does not prescribe how to conduct concrete perception tests. 
Instead, it specifies claims within a safety case for which manufacturers should deliver arguments and evidence. 
These claims also go into detail about sensing, perception, and machine learning~\cite[Sec. 8.3--8.5]{UL4600_voting_2019}. 
Manufacturers are given a certain freedom in how to provide evidence for their claims. 
For example, the claim that the environment perception can provide acceptable functional performance~\cite[Sec.~8.4.1]{UL4600_voting_2019} must be supported at least by the metrics false negative and false positive rate, and can be supported by ``other relevant metrics". 
In terms of existence uncertainty, false negatives (FNs) are objects missed by the SUT, whereas false positives (FPs) are wrongly detected ghost objects.
For completeness, true positives (TPs) are existing objects which are also perceived by the SUT.

\subsection{UNECE Regulation for Level 3 ALKS}
The UNECE regulation~\cite{UNECE2020lanekeepinglevel3} provides a regulatory framework for automated lane keeping systems (ALKS) of SAE Level 3 on highways. 
It describes the role of a technical service to whom the manufacturer has to demonstrate its compliance with the framework.
Stated requirements related to OEDR are a minimum field of view (FOV) of the perception subsystem and the ability to detect conditions that impair the FOV range \cite[Sec.~7]{UNECE2020lanekeepinglevel3}.
A minimum set of scenarios for testing the FOV range is specified verbally in \cite[Annex~5]{UNECE2020lanekeepinglevel3}. 
For instance, the perception of small targets like pedestrians or powered two-wheelers should be tested near the edges of the minimum FOV.
The technical service is supposed to select and vary the concrete parameters of these test scenarios.
Furthermore, the scenario parameters for system-level collision avoidance tests include also the perception-related conditions of the roadway and of lighting and weather \cite[Annex~4, Appendix~3]{UNECE2020lanekeepinglevel3}. 
Even though this regulation provides more specific test scenarios than other standardization approaches, they are not explicitly targeted to perceptual uncertainty, and also lack detailed specifications of for example environmental conditions. 

\subsection{Safety First for Automated Driving}
\label{sec:standards_safetyfirst}

This white paper by a consortium of startups, OEMs, and suppliers intends to contribute to an industry-wide safety standard for AD by expanding the considerations of ISO/PAS 21448 to SAE Levels~3 and~4~\cite{SaFAD2019}.

Related to this review, \cite[Sec.~3.3--3.6]{SaFAD2019} highlights that testing is essential, but not sufficient to assure the safety of AVs. %
It suggests a test methodology that includes decomposing the AD system into elements such as the perception subsystem, and testing these elements separately. 
For the statistical validation of the perception subsystem in real-world tests, the usage of a reference perception system 
and the so-called scenario-based approach (see e.g. \cite{Riedmaier2020_surveySBT}) are emphasized. 
Recorded raw data from the field should be re-processed offline upon perception algorithm updates. %
The scenario-based approach can argue about a sufficient coverage of relevant traffic scenarios by grouping recorded data that include certain influencing factors into equivalence classes of scenarios (more about interpreting recorded data as logical scenarios in Sec.~\ref{sec:axis_test_scenarios}).
Four constraints regarding perception testing are outlined \cite[Sec.~3.6.3]{SaFAD2019}:
\begin{itemize}
    \item The re-processing environment of the field recordings must be validated in terms of hardware and software.
    \item The test scenario catalog must be both statistically significant and covering the ODD sufficiently (see also Sec.~\ref{sec:describing_scenarios_odd},~\ref{sec:scenarios_obtaining}).
    \item The reference data quality must be appropriate for the validation objective (see also Sec.~\ref{sec:ref_data_tradeoffs}).
    \item Concrete test scenarios or data must be separate from those used in development (see also Sec.~\ref{sec:scenarios_different_sets}).
\end{itemize}

Additionally, \cite[Appendix~B]{SaFAD2019} explicitly addresses safety challenges regarding machine learning, which is omnipresent in an AV's perception, but has not yet been sufficiently addressed in ISO~26262 and ISO/PAS~21448.

\subsection{ISO/TR 4804}

The technical report ``Road vehicles - Safety and cybersecurity for automated driving systems - Design, verification and validation"~\cite{ISO_TR_4804_2020} aims at supplementing existing standards and documents on a more technical level.
Since its content is closely related to the previously published ``Safety First for Automated Driving" white paper~\cite{SaFAD2019}, we omit a repetition of the key aspects (Sec.~\ref{sec:standards_safetyfirst}).

\subsection{Conclusion of Safety Standards}

Contemporary automotive safety standards require arguments about successful OEDR for safety assurance, but mostly leave the realization of perception tests open to the AV manufacturer. 
This motivates the subsequent review of how perception testing is currently realized, and how it could be performed in future to assess if the perception allows safe vehicle behavior. 

\section{Established Activities of Perception Testing}
\label{sec:established}

This section expresses perception algorithm benchmarking outside the safety domain (Sec.~\ref{sec:developer_testing}) and data-driven sensor modeling on object-level (Sec.~\ref{sec:sensor_modeling}) in terms of the testing taxonomy~\cite{stellet2015testing}. 
We include both established activities into this separate section because they only partially target this paper's research question.
Thereby, we aim at providing a complete context for AV perception testing while distinguishing between directly safety-relevant and less safety-relevant testing activities. 

\subsection{Perception Algorithm Benchmarking}
\label{sec:developer_testing}

Developers of computer vision and moving object tracking algorithms across multiple industries are already using established benchmark datasets and test metrics for their algorithm's results. 
The idea behind these testing activities is typically to provide a quantitative ranking among different algorithms rather than evaluating whether safety-relevant pass-fail criteria are met.
The hardware and software that provide the raw sensor data %
are usually not analyzed.

\begin{figure}[t]
    \centering
    \vspace*{2mm}
    \includegraphics[width=\linewidth]{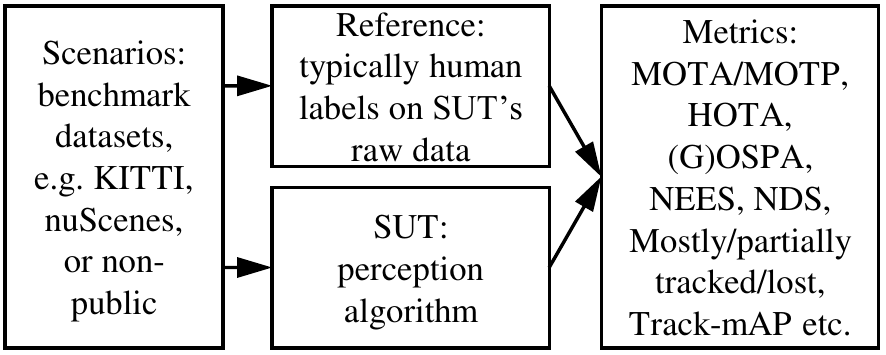}
    \caption{Perception algorithm benchmarking expressed in taxonomy (Fig.~\ref{fig:taxonomy_general}).}
    \label{fig:taxonomy_dev}
\end{figure}

In public research, the providers of public benchmark datasets typically determine all of the three testing axes test scenarios, reference data, and metrics (Fig.~\ref{fig:taxonomy_dev}). Test criteria usually mean that novel algorithms should rank higher than previous algorithms.

\subsubsection{Scenarios}
Test scenarios are captured when the dataset provider records the raw sensor data for the benchmark.
The aim for such raw data is typically not to capture safety-relevant scenarios, but rather to provide a diverse set of road user types that the algorithms under test will have to detect and distinguish. 
Therefore, for the purposes of perception algorithm benchmarking so far, it is sufficient that the recorded scenarios are only characterized in the data format of the recordings, and not in a dedicated scenario description language (Sec.~\ref{sec:describing_scenarios_odd}).
Examples for such datasets are KITTI~\cite{Geiger2012CVPR}, nuScenes~\cite{caesar2019nuscenes}, the Waymo Open Dataset~\cite{sun2019waymoopendataset}, or Argoverse~\cite{Chang2019Argoverse3T}. 

\subsubsection{Reference Data}
Besides raw sensor data, algorithm benchmarking datasets typically also provide reference data on object level in the form of bounding boxes that are labeled offline with the help of humans, e.g.~\cite{Geiger2012CVPR, caesar2019nuscenes, sun2019waymoopendataset}.  
For testing the world model output of the perception subsystem, reference data in a metric coordinate system over the ground plane are relevant, which can be labeled based on dense lidar point clouds.
In contrast, pixel-level class labels in camera images do not provide a reference for the resulting world model.

Other developer testing activities provide the reference information of target vehicles by means of high-precision global navigation satellite systems (GNSS), e.g.~\cite{stampfle2005performance}. 

\subsubsection{Metrics}
\label{sec:perc_algo_benchm_metrics}
After a perception algorithm has produced an object-based world model from the provided raw sensor data, its tracks can be compared to the provided reference tracks using metrics. 
Metrics that assess untracked object lists %
are also widely used, but are not discussed here because tracked object lists seem to be more common at the interface between the perception and planning subsystems.
Since a common goal of object tracking metrics is to enable a ranking of algorithms, they often need to summarize various aspects of the perception performance into a small number of scalar values that are used for comparison. 
This concept of aggregating various detailed low-level metrics into fewer summarizing high-level metrics has already been proposed in 2005 for the performance evaluation of automotive data fusion~\cite{stampfle2005performance}.

Computing lower-level metrics such as the rates of FPs and FNs requires to \textit{associate} (or \textit{match}, often used synonymously) the estimated tracks of the SUT to their corresponding reference tracks, which is realized in the following way. 
The pairwise distances between all estimated and all reference objects or tracks are computed by means of an object distance function. 
Examples for such distance functions are the Euclidean distance between object centroids in the ground plane (applied in e.g.~\cite{caesar2019nuscenes, sun2019waymoopendataset, Chang2019Argoverse3T}) or the Intersection over Union (IoU) of bounding box areas or volumes (e.g.~\cite{Geiger2012CVPR}). 
Mathematically speaking, however, IoU is not a proper distance function because it is bounded between 0 and 1 and only its complementary value ($1-IoU$) expresses distance. 

Using these pairwise distances, a multi-object association algorithm can compute an optimal association such that the sum of distances of all associated object pairs is minimized. 
Often, a threshold distance is applied to only allow reasonable object associations. 
For example, two objects with a Euclidean distance larger than, e.g., $\SI{2}{\metre}$~\cite{caesar2019nuscenes}, or with an IoU smaller than, e.g., $\SI{50}{\percent}$~\cite{Geiger2012CVPR}, could be prevented from becoming associated object pairs.
Examples for association algorithms include the Hungarian/Munkres algorithm~\cite{Kuhn1955_hungarianAlgo} and the auction algorithm~\cite{Bertsekas1989_auctionAlgo}. 

Once this association is obtained, different higher-level metrics can be computed to describe the overall perception performance. 
Since the last step of an overall perception algorithm is usually to track the frame-individual objects over time while potentially fusing them into a single list of tracks, the following metrics are often called multiple object tracking (MOT) metrics.

The most common MOT metrics are the CLEAR MOT metrics MOTA and MOTP, which stand for MOT accuracy and MOT precision, respectively~\cite{bernardin2008evaluating}. 
MOTA measures existence uncertainty by penalizing FPs, FNs, and ID switches, where ID switches are times when the ID of the estimated object changes while the ID of its associated reference object stays the same. 
MOTP measures state uncertainty by penalizing TPs that are not estimated precisely. 

Another state uncertainty metric that additionally measures whether an estimated object state is consistent with the tracker's estimate of its own Gaussian state uncertainty is the Normalized Estimation Error Squared (NEES, see e.g.~\cite{Chen_2018}).

The various forms and derivations of the Optimal Sub-Pattern Assignment (OSPA) metric~\cite{schuhmacher2008consistent,ristic2010performance,he2013track,vu2014new, beard2017ospa, rahmathullah2017generalized, shi2017comprehensive} penalize existence and state uncertainties at the same time by weighing them to obtain a single score. 
Unlike most other mentioned metrics, the OSPA metric is a metric in the mathematical sense, meaning it satisfies the three properties 1.) identity of indiscernibles, 2.) symmetry, and 3.) the triangle inequality.
This property makes it interesting for research on object tracking on a detailed mathematical level, e.g.~\cite{Reuter2014_LMB_paper, granstrom2016extended}.

Besides existence and state uncertainties, the nuScenes detection score (NDS)~\cite{caesar2019nuscenes} also considers classification uncertainties, as its goal is to represent the entire object tracking performance by only one scalar. 
Otherwise, metrics for classification uncertainty would be precision, recall, or the area under the so-called receiver-operating characteristic (ROC) curve~\cite{Fawcett2006ROC}.

One evaluation concept that explicitly considers temporal aspects of the object tracking performance is the Mostly tracked/Partially tracked/Mostly lost approach, which counts the number of reference trajectories that were tracked during more than $\SI{80}{\percent}$ or less than $\SI{20}{\percent}$ of their lifetimes~\cite{li2009learning}.

The so-called Higher-Order Tracking Accuracy (HOTA) metric has been proposed to balance various sub-metrics in a single higher-order metric~\cite{Luiten2020hota}. 
While the sub-metrics can describe individual performance aspects in detail, the higher-order metric can balance those sub-metrics without over-emphasizing one aspect over another~\cite{Luiten2020hota}.
We refer the reader to the same publication~\cite{Luiten2020hota} for a more general introduction to MOT algorithm benchmarking metrics and an in-depth analysis of popular metrics, which further include the mean average precision for tracking (Track-mAP) and the IDF1-score. 

Recently, the ``Planning Kullback-Leibler divergence" (PKL)~\cite{Philion2020planner_centric} has emerged from the field of perception algorithm benchmarking. 
It is the only metric from this field that we are aware of that explicitly considers the actual influence of perceptual uncertainties on the downstream motion planner. 
Various influences on the metric have been analyzed on the submissions of the nuScenes object detection challenge~\cite{Guo2020efficacy}. 
Interestingly, the submission rankings would be significantly different if PKL was used as the main benchmark instead of mAP~\cite{Guo2020efficacy}. 
Moreover, the PKL was found to be more consistent with human intuition than the NDS about which perceptual uncertainties are actually dangerous~\cite{Philion2020planner_centric} and has been included into the official submission evaluation.
Due to its direct relevance for this paper's research question, technical details on the PKL are discussed later along with other safety-aware microscopic metrics (Sec.~\ref{sec:downstream_comparison}).

\subsubsection{Difficulties with Association Uncertainty}
\label{sec:association_uncertainty}

There seems to be a fuzzy border between \textit{state} and \textit{existence} uncertainties, because a fixed threshold on the distance function that distinguishes TPs from FPs/FNs is not likely to produce intuitive associations under all circumstances. 
Therefore, the studies~\cite{Florbaeck2016.matching.offline} and~\cite{Sondell2018} tune their offline object association in a way that it reproduces human annotations of object association. 
These approaches seem to lead to subjectively better associations, but have the cost of less human understanding of how the trained associator and subsequently also the test metric works.

\subsubsection{Relevance for Vehicle Safety}
\label{sec:safety_relevance_dev_metrics}

Except for PKL~\cite{Philion2020planner_centric}, the mentioned state-of-the art metrics for object perception and tracking do not consider safety, but rather provide information about the average similarity to a reference dataset~\cite[Safety Concern~9]{Willers2020safety}. 
This issue is described in more detail in~\cite{Piazzoni2020Modeling_conference}.
According to~\cite{Salay2019partialspecifications}, it is possible to completely specify pass-fail criteria for such safety-independent metrics. 
Also ISO/PAS 21448 suggests to use safety-independent perception metrics \cite[Annex~D, 11)]{iso2019sotif}. 
However, those criteria would only represent non-functional requirements, while functional requirements needed for safety assurance remain an open issue~\cite{Salay2019partialspecifications}.

Similarly, the authors of~\cite{Aravantinos2020} point out that metrics like a FP rate cannot provide safety-relevant information because some FPs might be highly safety-relevant while others are not. 
Therefore, they suggest to formulate realistic fault models, which already exist in ISO 26262 for e.g. hardware faults, also for perception algorithms. 
Such fault models should depend on the safety requirements of the overall vehicle.
Given examples include the differentiation of e.g. a FP for a pedestrian from a FP for a bicycle, or to discretize continuous state uncertainty to obtain boolean faults.
Nevertheless, formulating such fault models in accordance with the safety requirements seems non-trivial, which motivates for a further analysis of metrics and test criteria in Sec.~\ref{sec:axis_criteria_metrics}.

\subsection{Object-Level Data-Driven Sensor Modeling}
\label{sec:sensor_modeling}

\begin{figure}[t]
    \centering
    \vspace*{2mm}
    \includegraphics[width=\linewidth]{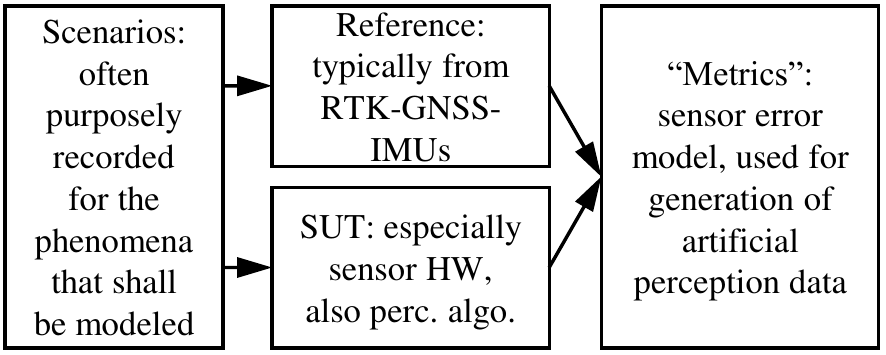}
    \caption{Object-level data-driven sensor modeling expressed in taxonomy (Fig.~\ref{fig:taxonomy_general}).}
    \label{fig:taxonomy_sensor_modeling}
\end{figure}

The idea of data-driven sensor modeling approaches is usually to treat the complex behavior of a sensor as a black box and replace it by a model which can generate artificial sensor data in simulations. 
To do so, the output of a sensor under test is compared to reference data in order to train or parametrize a model that describes the sensor's perception performance.
Hence, sensor modeling can also be interpreted as a testing activity according to the taxonomy of this paper (Fig.~\ref{fig:taxonomy_sensor_modeling}).

\subsubsection{System Under Test/System to be Modeled}
In the context of sensor modeling, we use the terms \textit{system under test} and \textit{system to be modeled} synonymously. 
Similar to the research interest of this paper, sensor modeling approaches like~\cite{hirsenkorn2015nonparametric, hanke2016classification, Krajewski2020UsingDrones, Zec2018markovModeling} model object lists, which have been generated through both sensor hardware and perception software. 
This is common if the sensor's built-in perception algorithm is inaccessible to the modeling engineer due to sensor supplier IP. %
Otherwise, the virtually tested AD function could contain parts of the perception algorithm, reducing the system to be modeled to mostly the sensor hardware~\cite{holder2019modeling}. 
However, modeling the already processed perception data has the advantages of less data volume and more comparable data formats. 

\subsubsection{Scenarios}
Like in Perception Algorithm Benchmarking (Sec.~\ref{sec:developer_testing}), test scenarios for the SUT are captured during data recording. 
 Already seemingly simple test scenarios with dry and sunny weather conditions and with only few other road users often provide enough difficulty for data-driven modeling of the sensor hardware behavior. 
 For example, modeling radar reflections on guardrails on a motorway is a topic of current research~\cite{holder2019modeling}.
 Unlike for perception software development, the research field of sensor modeling seems to be lacking common benchmarking datasets that would allow a straightforward comparison of different sensor models on the same data.

\subsubsection{Reference Data}
Modeling the behavior of sensor hardware in a data-driven approach requires an independent reference to the sensor hardware. 
This makes RTK-GNSS-IMUs (see Sec.~\ref{sec:ref_data_rtk_gnss_imu}) a suitable and typically used source of reference data (see e.g.~\cite{schaermann2017validation, hirsenkorn2015nonparametric} for their application).

\subsubsection{Metrics}

Metrics in this paper are quantitative statements on the SUT.  
Since parametrized or trained sensor models can describe how similar the SUT's perception is to the reference perception, they are interpreted as test metrics here. 
Note that sensor models are usually not single-score metrics, but rather for example probability distributions of the sensor's errors~\cite{hirsenkorn2015nonparametric}. 
Note also that the term \textit{metric} in sensor modeling literature is used differently - there, metrics make a statement on the modeling approach rather than on the SUT 
\cite{rosenberger2019towards}.

Actual statements on the SUT can be for example the mean values and standard deviations of Gaussian distributions that describe the SUT objects' state errors in e.g. position and velocity~\cite{Krajewski2020UsingDrones}. 
A nonparametric distribution for such errors that is based on a Gaussian mixture model is used e.g. in~\cite{hirsenkorn2015nonparametric}.
There are various further ways of expressing the SUT's errors by sensor models, which are however mostly outside the focus of this paper.

\subsubsection{Relevance for Safety-Aware Perception Testing} %

Typically, sensor models aim at describing the specific phenomena of a sensor modality as detailed as feasible, no matter how safety-relevant these phenomena are~\cite{hanke2016classification, rosenberger2019towards}.
However, some sensor modeling activities explicitly address this paper's research question.
For example,~\cite{Piazzoni2020Modeling_conference} models perception errors while considering the effect they have on robust decision making. 
In the context of validation of sensor models,~\cite{Holder2020radarModelPlanningInfluence} argues that a key property of modeled sensor data should be that they induce the same behavior in the downstream driving function like the real sensor data would do. 
Such topics are further elaborated in Sec.~\ref{sec:metrics_perc_control_linkage} about the perception-control linkage.

\section{Test Criteria and Metrics}
\label{sec:axis_criteria_metrics}

This section and the following two sections each cover one testing axis of the used taxonomy from~\cite{stellet2015testing} (Fig.~\ref{fig:taxonomy_general}) and are dedicated to the primary research question of this paper.

According to the taxonomy, test criteria and metrics are ``a statement on the system-under-test (test criteria) that is expressed quantitatively (metric)"~\cite{stellet2015testing}.
For example, a criterion %
that is qualitative at first could be quantified by means of specifying intervals on a related metric that determine passing or failing a test.

After examining how to specify perception requirements and test criteria (Sec.~\ref{sec:requirements}), we deal with safety-aware microscopic metrics and criteria (Sec.~\ref{sec:safety_metrics_micro}). 
Furthermore, metrics on the self-reporting and confidence estimation capabilities of the SUT are discussed (Sec.~\ref{sec:self_reporting_metrics}), as well as macroscopic metrics towards approval (Sec.~\ref{sec:safety_metrics_macro}).

\subsection{Specification of Requirements and Criteria}
\label{sec:requirements}

A perception subsystem shall enable the AV to reach its destination safely, comfortably, and in reasonable time. 
Therefore, it shall provide information in sufficient quality about all road users that are relevant for fulfilling the driving task.
However, such requirements are not specific enough to be tested~\cite{Philipp2020DecompositionPerception}.
Thus, how can one specify ``relevant for fulfilling the driving task", or ``sufficient quality" for usage as binary test criteria?

In the following, it may be useful to differentiate between concept specification and performance specification of environment perception~\cite{Czarnecki_2018_Framework}. 
In the mentioned source, concept specification refers to defining the properties to be perceived such as a pedestrian's pose, extent, and dynamic state, given an ODD. 
In contrast, the performance specification defines how well these properties should be perceived, for example in terms of detection range, confidence, and timing~\cite{Czarnecki_2018_Framework}. 

\subsubsection{The Difficulty of Specifying Perception}

AVs are expected to operate in unstructured, public, real-world environments, which are called \textit{open context} in~\cite{poddey2019opencontext}.  
According to~\cite{Spanfelner2012ChallengesISO26262}, a complete concept specification of the environment perception may not be possible because a model about the environment generally cannot cover all necessary relations and properties in such an open context. 
This issue has also been called ontological uncertainty~\cite{Gansch2020uncertainty}.
For example, to specify the perception of pedestrians, one would have to specify what a pedestrian is, which is, however, only partially possible using rules such as necessary or sufficient conditions~\cite{Salay2019partialspecifications}.
Providing examples of pedestrians in a training set is how machine learning engineers specify the concept of a pedestrian. 
On the one hand, this can enable driving automation of SAE Levels 3 and above, but on the other hand, it prevents a traditional specification according to ISO 26262~\cite{Salay2019partialspecifications}.

Besides specifying the perception concept and performance for %
discrete environmental aspects like the classification of a road user, another key challenge is to identify when uncertainty in continuous and dynamic environmental aspects, like a car's velocity, leads to safety-relevant failures~\cite{Philipp2020DecompositionPerception}.

\subsubsection{Concrete Approaches of Specifying Perception} 

Contributions from the field of machine learning investigate how to specify the perception subsystem by using pedestrian detection as a benchmark example~\cite{Salay2019partialspecifications},~\cite{Rahimi2019requirements}. 
On a higher level,~\cite{Salay2019partialspecifications} emphasizes the importance of an adequate language for specification and the potential of deliberate \textit{partial} specifications. 
The paper proposes and evaluates several methods for incorporating partial specifications into the development process.
Further literature with concrete specification approaches include creating an ontology of the exemplary ``pedestrian" domain (concept specification)~\cite{Rahimi2019requirements}, and taking human perception performance as a reference (performance specification)~\cite{Hu2020}.
A formal language for specifying requirements on the performance of object detection in the absence of reference data is proposed in \cite{Dokhanchi2018qtl,Balakrishnan2019metrics}.

Instead of using an environmental concept like a pedestrian as the center of investigation, the methodology of~\cite{Klamann2019} outlines how particular test criteria for AV subsystems can be defined in a top-down way, starting from overall safety goals. 
The methodology is, however, not yet applied specifically to the perception subsystem. 

As mentioned earlier, the complexity of the open context can cause gaps in the specified requirements. 
To fill these gaps, many organizations collect large numbers of real-world mileage to discover so far unknown scenarios~\cite{Koopman2018towardFramework}.   
These recorded scenarios can also serve as test scenarios if they are identified as test-worthy (Sec.~\ref{sec:scenarios_obtaining}).

Besides the approaches mentioned so far, there are also considerations on perception requirements that aim at complying with the traditional ISO 26262 functional safety standard. 
The sources~\cite{Cassel2020SAEPerceptionRequirements,Johansson2016perceptionASIL,Johansson2017assessingUseOfSensors} propose dynamically associating an Automotive Safety Integrity Level (ASIL) with a given driving situation such that for example, the perception is required to comply with the stricter ASIL D in high-risk situations and with the less strict ASIL A in low-risk situations.
An example of how functional safety requirements for the perception subsystem can be derived based on a fault tree analysis (FTA) is given in~\cite{schoenemann2019fault} in the context of automated valet parking.

The following subsections deal with metrics for measuring quantitatively whether specified test criteria are met. 

\begin{figure*}[t]
    \centering
    \vspace*{2mm}
    \includegraphics[width=\textwidth]{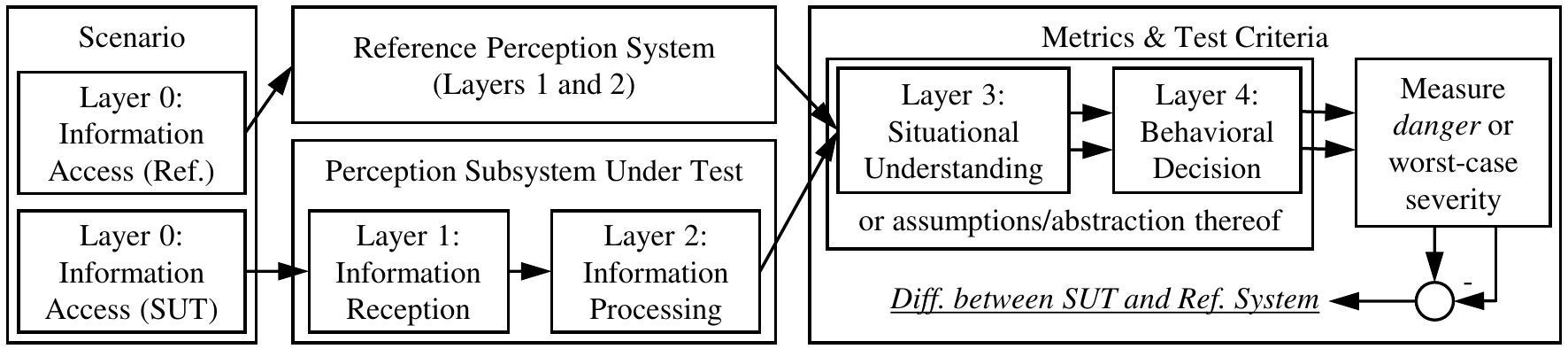}

    \caption{Safety-aware perception testing of~\cite{shalevshwartz2017formalRSS} and~\cite{Salay2019_safety_perceptual_components} expressed in taxonomy for testing by~\cite{stellet2015testing} (Fig.~\ref{fig:taxonomy_general}), with decomposition layers of the driving function by~\cite{amersbach2017functional} (Fig.~\ref{fig:amersbach_layers}) mapped into the taxonomy.
    Further explanation is given in Sec.~\ref{sec:downstream_comparison}.}
    \label{fig:taxonomy_amersbach_layers}
\end{figure*}

\subsection{Microscopic Test Criteria and Metrics} %
\label{sec:safety_metrics_micro}

The currently most established perception performance metrics, which are typically used in machine learning, do not represent whether the perception output is sufficient for safe vehicle operation \cite[Safety Concern~9]{Willers2020safety} (Sec.~\ref{sec:safety_relevance_dev_metrics}). 
This section therefore reviews literature about metrics and criteria that explicitly distinguish safety-relevant from safety-irrelevant perception errors.

\subsubsection{Heuristic for the Safety-Relevance}
\label{sec:heuristic_safety_relevance}

The authors of~\cite{Berk2020summary} provide a simple approach towards the safety-criticality of perception errors in the existence uncertainty domain. 
Certain fractions of the binary error types FP and FN are assumed to be safety-critical, where the fractions depend on the error's position of occurrence within the ego vehicle's FOV. 
For example, perception errors directly in front of the ego vehicle can heuristically be estimated to be more likely to be safety-critical than perception errors occurring farther away or with a lateral offset~\cite[Figure~3]{Berk2020summary}. 
The benefit of this approach is that once its numerical values are set, it does not need to consider any downstream driving function for computing safety-critical failure rates. 

\subsubsection{Modeling the Perception-Control Linkage}
\label{sec:metrics_perc_control_linkage}
However, whether a perception error turns out to be safety-critical or not generally does depend on the downstream driving function. 
Thus, we describe different ways of modeling the interface between perception and planning/control, which is also called the perception-control linkage~\cite{Salay2019_safety_perceptual_components}. 

The above-mentioned fractions of safety-critical perception errors could be determined by means of closed-loop fault injection simulations~\cite{Berk2020summary}. 
However, if the downstream driving function receives a minor update, the perception metrics would have to be recomputed, which would render such a direct approach unpractical for iterative development. 

Alternatively, one could abstract the perception and planning subsystems such that test results of the perception subsystem can be re-used for varying planners. 
For this purpose, modular functional system architectures~\cite{amersbach2017functional},~\cite{Philipp2020DecompositionPerception}, \cite{ulbrich2017functional},~\cite{tas2016functional}
could be implemented with contracts, assumptions, and guarantees at the interfaces between the perception and planning subsystems~\cite{Burton2017SafetyML_HAD},~\cite{Burton2019confidence_arguments},~\cite{Meyer1992designByContract},~\cite{Cassel2020SAEPerceptionRequirements}.

Such a modular approach would enable different safety assurance methodologies for the different subsystems. 
For example, the following section discusses data-driven perception testing and formal safety assurance of the planner. 
The idea behind such formal methods is to always assure a certain safe planning behavior, given that the environment is perceived well enough. 
Popular formal models for safe planning include Responsibility-Sensitive Safety (RSS)~\cite{shalevshwartz2017formalRSS} and reachability analysis~\cite{Pek2020onlineVerification, althoff2010reachability}. 

\subsubsection{Downstream Comparison of SUT and Reference Data}
\label{sec:downstream_comparison}

Traditionally, perception metrics directly compare the object list of the SUT to the one of the reference system. 
With an additionally given planner or assumptions about it, safety-aware perception metrics can be computed indirectly. 
The planner, or a formal abstraction thereof, computes behavior outputs for world model inputs from the SUT and from the reference system. 
Safety-aware metrics then penalize if the behavior induced by the SUT is different from the one induced by the reference system, especially in terms of danger and worst-case severity (Fig.~\ref{fig:taxonomy_amersbach_layers}). 

The previously mentioned PKL metric~\cite{Philion2020planner_centric, Guo2020efficacy} (Sec.~\ref{sec:perc_algo_benchm_metrics}) works is this fashion.
It uses an end-to-end machine-learned planner, which is trained to imitate the human ego vehicle driving behavior of the recording of the dataset on which it will be applied. 
During inference, this planner is fed with the reference object list and with the object list of the SUT. 
For both object lists in a given scene, it computes the likelihoods of the recorded ego vehicle trajectory during a short time interval after the given scene. 
A ``Planning Kullback-Leibler divergence" (PKL) between these likelihoods then represents the degree to which the SUT's uncertainties influence the planner.

While PKL uses a concrete planner, the RSS framework~\cite{shalevshwartz2017formalRSS} defines safety-relevant perception failures using a formal abstraction of the planner and its formal definition of \textit{dangerous} situations. %
According to the publication, \textit{safety-critical ghosts} are situations in which the environment perceived by the SUT is formally considered \textit{dangerous} for the ego vehicle even though according to the reference data, it is not. 
Likewise, \textit{safety-critical misses} are situations in which the environment perceived by the SUT is formally not considered \textit{dangerous} even though it actually is. 
For example, a safety-critical ghost can be a false-positive pedestrian detection in front of the ego vehicle that could cause a false and dangerous braking maneuver. 
A safety-critical miss could be a pedestrian at the same location as a false-negative detection, which could lead to the ego vehicle hitting the pedestrian. 
Safety-critical ghosts and misses can not only be caused by existence uncertainties, but also by state and classification uncertainties~\cite{shalevshwartz2017formalRSS} because their definition is agnostic of the type of uncertainty. 
This avoids a potentially ambiguous differentiation between state and existence uncertainties (see Sec.~\ref{sec:association_uncertainty}).

However, this concept of safety-critical perception failures~\cite{shalevshwartz2017formalRSS} so far seems to have been treated mostly theoretically. 
For example, we are not aware of its extension to consider multiple time steps %
or its practical demonstration on real data. 

Salay et al.~\cite{Salay2019_safety_perceptual_components} expand this concept of binary safety-critical perception failures to a potentially continuous description of the \textit{risk} that a perception failure can cause. 
The work formalizes a concept to analyze the so-called \textit{incurred severity} %
that an uncertain environment perception can cause. 
It also requires knowledge about the planning subsystem, for example, the planner's policy for computing actions based on a world model. %
Furthermore, for actions that can cause harm, there must be a way to assess the worst-case severity of this harm. 
With these assumptions, the microscopic and safety-aware perception metric \textit{incurred severity} is defined as the difference between the SUT's and the reference system's induced worst-case severities of control actions~\cite{Salay2019_safety_perceptual_components} (Fig.~\ref{fig:taxonomy_amersbach_layers}). 

The same work~\cite{Salay2019_safety_perceptual_components} applies this concept in a case study dealing with classification uncertainty of road users. 
For example, if a pedestrian is correctly perceived as a pedestrian, then the worst-case \textit{incurred severity} is zero. 
If a pedestrian is however classified as a cyclist, and if the control action for cyclists is less cautious than for pedestrians, then the \textit{incurred severity} of this misclassification is likely positive. 
Future work not yet covered in~\cite{Salay2019_safety_perceptual_components}  
is to generalize the computation of \textit{incurred severity} to world models that are also subject to existence and state uncertainty. 
Moreover, representative real-world exposures (probabilities of occurrence) of scenes would be needed for valid risk computations.

\begin{figure}[t]
    \centering
    \vspace*{2mm}
    \includegraphics[width=\linewidth]{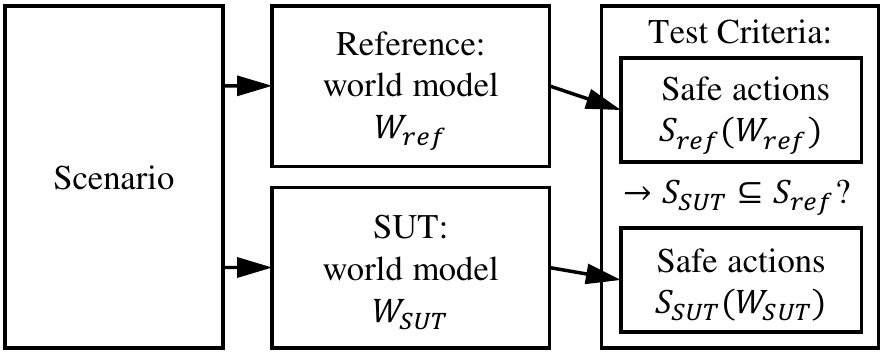}

    \caption{%
    Testing whether the SUT's world model can induce control actions that would not be safe for the reference world model~\cite[Def.~2.1]{Salay2020_purss}.
    Figure follows taxonomy of Fig.~\ref{fig:taxonomy_general}. 
    }
    \label{fig:taxonomy_purss}
\end{figure}

A practical challenge in the approach of \cite{Salay2019_safety_perceptual_components} might be the availability of the planner's behavior policy. 
This challenge is addressed in the subsequent positional paper by Salay et al.~\cite{Salay2020_purss} on perceptual uncertainty-aware RSS (PURSS). 
It proposes to let the RSS framework handle world model uncertainties in a more advanced and more practical way than originally published. 
A behavior policy of the concrete planner is not needed; instead, it is assumed that the planner complies with the formal RSS rules for guaranteed safe control actions. 
The RSS model, which is explained in detail in \cite{shalevshwartz2017formalRSS}, can provide a set of guaranteed safe control actions for any given world model input. 
Using this set, safety-aware test criteria for the perception subsystem can be defined. 
The set of safe control actions is computed for both the world model from the SUT and for the reference world model. 
The safe control actions $S_{SUT}$ for the SUT's world model is what a real planner would execute, whereas the safe control actions $S_{ref}$ for the reference system's world model is what is actually safe. 
If there are control actions in $S_{SUT}$ that are not part of $S_{ref}$, then the SUT potentially causes a safety risk \cite[Def.~2.1]{Salay2020_purss}  (Fig.~\ref{fig:taxonomy_purss}). 

Furthermore, the authors propose how RSS could be extended to handle imprecise world models, which are sets of SUT world models %
that contain the true world model with a certain confidence. %
For example, the position of a pedestrian could be estimated by an interval of $[\SI{29}{\metre}, \SI{31}{\metre}]$ with a confidence of $\SI{95}{\percent}$. 
In practice, contemporary perception algorithms do output confidences about their world models, for example Gaussians for state uncertainties, which could be sampled at the mentioned confidence intervals.
A corresponding test criterion for an imprecise world model is that its RSS-defined set of safe control actions must be safe for each precise world model contained in it \cite[(2)]{Salay2020_purss}. 
The authors show that the mentioned perception confidence can be used as a lower bound for the probability of guaranteed safe control actions~\cite[(3)]{Salay2020_purss}. %

\subsection{Metrics for Uncertainty/Confidence Calibration}
\label{sec:self_reporting_metrics}

As described in the previous section and also in~\cite{Koopman2018towardFramework}, the explicit consideration of world model uncertainty (or inversely, confidence) can facilitate safety assurance, for example through runtime monitoring (more in Sec.~\ref{sec:uncertainty_tolerance}).
In this context, it is crucial that the self-reported uncertainty of the SUT correctly reflects its true uncertainty~\cite[Safety Concern~5]{Willers2020safety}.
If this holds, then the SUT is \textit{calibrated}~\cite{Guo2017calibration}, which means that it is neither over-confident nor too doubtful of itself.

\subsubsection{Types of Uncertainty}
\label{sec:uncertainty_forms}

The literature on uncertainty in AD (e.g.~\cite{Arnez2020_uq_comparison}) describes that \textit{epistemic} uncertainty, or model uncertainty, constitutes how uncertain the SUT's model of deep neural networks (DNNs) is in correctly describing the environment. 
In contrast, \textit{aleatoric} uncertainty is caused by physical sensor properties like finite fields of views, resolutions, and sensor noise. 
Both types are relevant for correct calibration.
Additionally, \cite{Gansch2020uncertainty} proposes the concept of \textit{ontological} uncertainty to describe the complete unawareness of certain aspects of the environment, even in the reference data. 

\subsubsection{Representations of Uncertainty}
\label{sec:uncertainty_representations}

In object perception, state estimation is a regression task (continuous true value), while existence and classification estimation are classification tasks (discrete true values)~\cite{Arnez2020_uq_comparison}. %
State uncertainties are typically quantified by continuous probability distributions or confidence intervals, 
while existence and classification uncertainties are usually expressed by scalar probabilities between $0$ and $1$. 

\subsubsection{Calibration Metrics}
\label{sec:uncertainty_metrics_detail}

This section focuses on metrics that only describe how well-calibrated the world model uncertainties are. %
In contrast, uncertainty-aware classification or regression metrics~\cite{Arnez2020_uq_comparison} are a topic of perception algorithm benchmarking (Sec.~\ref{sec:developer_testing}).

Multiple literature sources set up so-called calibration curves~\cite{Naeini2015calibrationError, Kuleshov2018regressionUncertainty, Guo2017calibration} for visualization and numerical analysis. 
These curves plot the accuracy (or empirical frequency) of a prediction over the prediction's confidence, often using bins. 
For example, in classification tasks, one specific bin could contain all events where the detector reports a confidence/existence probability in between $\SI{80}{\percent}$ and $\SI{90}{\percent}$. 
If there is a pedestrian present for only $\SI{70}{\percent}$ of all events in that bin, then the calibration curve would be tilted off from the ideal diagonal line. 
In regression tasks, a similar calibration curve can be set up by defining the accuracy as the empirical frequency of when the estimated confidence interval contained the true value.

Single-score calibration metrics can be computed from such calibration curves~\cite{Arnez2020_uq_comparison}. 
In classification tasks, the \textit{Expected Calibration Error} (ECE) and the \textit{Maximum Calibration Error} (MCE)~\cite{Naeini2015calibrationError} represent the expected and the maximum difference between confidence and accuracy in such calibration curves, respectively.
Similar metrics for regression tasks are the \textit{calibration error}~\cite{Kuleshov2018regressionUncertainty} or the related \textit{Area Under the Calibration Error Curve} (AUCE)~\cite{Gustafsson2020calibrationError}. 

Concrete applications of uncertainty calibration and evaluation in 3D object detection from lidar point clouds are given for example in~\cite{Feng2018uncertainty} and the follow-up publication~\cite{Feng2019calibrationECE}.
The latter publication explicitly uses the mentioned calibration curves and the ECE metric. 

A further uncertainty evaluation approach that explicitly addresses perception systems in safety-relevant domains is provided in~\cite{henne2020benchmarking} and subsequently~\cite{Schwaiger2020uncertainty}.
The publications, which focus on out-of-distribution detection for classification tasks, distinguish four different cases in classification results. 
The cases are combinations of the properties certain/uncertain (decided by a threshold), and correct/incorrect. 
Metrics are defined based on the fractions of these individual cases among all classification results. 
The fraction of certain, but incorrect results is defined as the \textit{Remaining Error Rate} (RER), and the fraction of certain and correct results is defined as the \textit{Remaining Accuracy Rate} (RAR). 

All mentioned metrics so far assume that a precise world model is available as a reference. 
However, also reference data are generally uncertain and imprecise (Sec.~\ref{sec:ref_data_uncertainty}), which makes test metrics more complex.

\subsection{Macroscopic Metrics Towards Approval} 
\label{sec:safety_metrics_macro}

So far in this paper, the used metrics and test criteria have been statements on the SUT in either individual scenes (microscopic metrics) or in relatively small amounts of test data with a research purpose.
While microscopic metrics and criteria are necessary for a detailed analysis of the SUT, they alone are not sufficient for the overall safety assurance of AVs because their results need to be extrapolated to estimate \textit{macroscopic}~\cite{Junietz2018criticalityMetric} metrics (definition in Sec.~\ref{sec:def_metric}). 

\subsubsection{Terminology: Safety vs. Reliability}
Since the term \textit{reliability} is often used in the context of macroscopic safety metrics~\cite{Berk2020summary}, it is first briefly differentiated from the term \textit{safety}. 
The perception subsystem of an AV must enable the overall vehicle to drive safely. 
Safety, however, is a property of the overall system and not of the perception subsystem. 
The perception subsystem can at best be \textit{reliable}, which is generally different from being \textit{safe}~\cite[Sec. 2.1]{leveson2016engineering} (Sec.~\ref{sec:definitions}). 
Note that a property that includes both safety and reliability is \textit{dependability}~\cite{Avizienis2004dependability}.

\subsubsection{Mean Time Between Failures}
The RSS publication~\cite{shalevshwartz2017formalRSS} argues that for the acceptance of AVs, the overall probability of occurrence of safety-critical failures must be in the order of magnitude of $10^{-9}$ per hour.
This would imply a test criterion of no such failures during about $\SI{e9}{}$ hours of driving, where safety-critical failures in their publication are the previously introduced safety-critical ghosts and misses. 
The related paper~\cite{Weast2020}, which focuses on perception within RSS, states another large required mean time between failures of $\SI{e7}{\hour}$ for the perception subsystem.
Only perception failures that can possibly lead to unsafe vehicle behavior should be counted in that number. 

Major contributions to the macroscopic reliability assessment of the perception subsystem have come from Berk et~al., e.g.~\cite{Berk2020summary},~\cite{Berk2019Dissertation}. 
The macroscopic metrics they use are the rate of all perception failures and the rate of safety-critical perception failures, both measured in failures per hour. 
Safety-critical perception failures are computed by filtering all perception failures with a field-of-view-dependent safety-criticality factor between $0$ and $1$, as explained in Sec.~\ref{sec:heuristic_safety_relevance}.
An overall approval criterion according to~\cite{Berk2019Dissertation} is that the sum of the failure rates of the perception, planning, and actuation subsystems is smaller than a given threshold rate, which is called the target level of safety.
A corresponding testing methodology to compute these macroscopic perception metrics is discussed in Sec.~\ref{sec:uncertainty_forecasting}.

\subsection{Conclusion of Test Criteria and Metrics}

Different metrics describe different properties of the SUT like its safety-relevant performance in a single scene, the calibration of its self-reported uncertainties, or its statistical safety impact.
Specifically the representation of safety-relevance in microscopic metrics seems challenging. 
These mentioned different classes of metrics are typically coupled. 
For example, the SUT's statistical safety impact depends on its safety impact in individual scenes, which in turn may depend on whether the SUT has correctly reported its uncertainty in that scene. 
Thus, one challenge is to harmonize individually dedicated metrics into an overall framework.
Moreover, the specification of test criteria is nontrivial due to the complexity of the open-world context. 

\section{Test Scenarios}
\label{sec:axis_test_scenarios}

According to the taxonomy of~\cite{stellet2015testing}, a test scenario is ``a set of specified conditions" under which a test is executed. 
The following subsections deal with adapting the term \textit{scenario} to the perception context (Sec.~\ref{sec:what_are_perc_scenarios}), describing test scenarios and the ODD (Sec.~\ref{sec:describing_scenarios_odd}), generating a test scenario catalog (Sec.~\ref{sec:scenarios_obtaining}), executing scenarios as test cases (Sec.~\ref{sec:executing_scenarios}), and splitting scenarios into training and test sets (Sec.~\ref{sec:scenarios_different_sets}).
Much of the literature on scenario-based testing does not yet focus on perceptual uncertainty, which is why this section also cites related literature with a broader focus.

\subsection{Adapting the Term Scenario to the Perception Context}
\label{sec:what_are_perc_scenarios}
The definition of the term \textit{scenario} has been discussed extensively in the context of scenario-based testing %
\cite{Ulbrich2015}.
According to the mentioned source, a scenario contains actors (mostly road users) whose actions influence the temporal development of scenes within the scenario. 
For testing purposes, one of such actors within the scenario is the subject vehicle. 
So far, the focus of scenario-based testing of AVs has been on the plan and act subsystems instead of on the perception subsystem of the subject vehicle~\cite{Riedmaier2020_surveySBT}. 
However, if only the perception subsystem is tested in an open-loop fashion, then it typically only has the role of a passive observer of the scenario instead of an influencing actor. 
At this point, we exclude closed-loop testing because we are interested in the offline test of potentially multiple perception algorithms on the same recorded raw sensor data from the real world. 

To the best of our knowledge, we are not aware of literature that explicitly considers the influence that an open-loop perception subsystem may have on the temporal development of a scenario. %
Such an influence is in fact possible through e.g. electromagnetic waves emitted by active sensors, which could disturb the sensors of other AVs and thus theoretically influence their behavior.
However, in this review, we assume for simplicity that perception-specific scenarios are scenarios %
whose temporal development of objective scenes is pre-determined, but where the subjective observation of these scenes in world models is left open. %
This corresponds to \textit{static scenarios} with an additionally predetermined ego vehicle behavior according to~\cite{Rocklage2017scenarios}.

Furthermore, it seems that the different abstraction levels of scenarios (functional, logical, concrete) of~\cite{Menzel2018scenarios} can also be used for perception scenarios. 
A functional scenario is described verbally (e.g. it rains, among other aspects), whereas a logical scenario has parameters (e.g. uniform precipitation in millimeters per hour), and a concrete scenario gives each parameter a numerical value (e.g. $\SI{2}{mm \per \hour}$).
Generally, different logical scenario parameters are of interest for perception tests than for planning tests~\cite{Neurohr2021criticality}. 

\subsection{Description of Scenarios and ODD}
\label{sec:describing_scenarios_odd}

Arguing whether test scenarios cover the entire ODD requires ways to describe scenarios and the ODD.

Current ODD description approaches are the taxonomy of BSI PAS 1883 \cite{bsi2020ODDStandard}, the NHTSA framework~\cite[Ch.~3]{Thorn2018NHTSA}, and  the ongoing standardization of a machine-readable ODD format in ASAM OpenODD~\cite{Asam2020openODD}.
Furthermore, ontologies are used to describe ODDs~\cite{Czarnecki2018OntologyPart2} and scenes within scenarios~\cite{Bagschik2018_ontology}. 
Ontologies are intended to be standardized in ASAM OpenXOntology~\cite{Asam2020openXontology} and have also been proposed to define a schema for world model data~\cite{Salay2020_purss}.
For ontology-generated scenarios, the notion of \textit{abstract scenarios} was introduced to describe functional scenarios that are formal and machine-readable, but do not yet have a logical parameter space~\cite{Neurohr2021criticality}. 
The ontology of~\cite{Bagschik2018_ontology}, which is used to generate traffic scenes for scenario-based testing, organizes all scenario entities by means of a 5-layer-model. %
This model was first defined using four layers in~\cite{Schuldt2017diss}, and has been adapted to six layers in~\cite{Bock2018databasis} and most recently~\cite{Scholtes2020_6LM}. 
The model's individual layers according to the most recent and also summarizing publication~\cite{Scholtes2020_6LM} are 

\begin{enumerate}
    \item Road network and traffic guidance objects
    \item Roadside structures
    \item Temporary modifications of Layers 1 and 2
    \item Dynamic objects
    \item Environmental conditions
    \item Digital information.
\end{enumerate}

These scenario description layers should not be confused with the functional decomposition layers of~\cite{amersbach2017functional} (Fig.~\ref{fig:amersbach_layers}).
Even though the present review only focuses on perceiving dynamic objects in Layer 4, the properties of all other layers can influence the perception of dynamic objects and are therefore also relevant.

So far, the layer models for scenario description only provide a way for an objective description of the environment within a traffic scenario. 
Aspects related to a subjective perception of such scenarios through machine perception are not yet explicitly covered, for example surface materials that affect radar reflections~\cite{Scholtes2020_6LM}. 
Such shortcomings of scenario descriptions for perceptual aspects have also already been identified in the context of sensor modeling~\cite[Sec.~V]{holder2019modeling}.

The previously mentioned scenario description methods are suited to describe functional, abstract, and perhaps logical scenarios. 
In contrast, describing concrete scenarios for the actual test execution requires a certain data format. 
The ASAM OSI~\cite{Driesten2019osi} classes GroundTruth and SensorView appear to be capable of describing objective scenes and the circumstances for their subjective perception, respectively. 
However, they are designed for virtual simulations instead of real-world test scenarios.
Similarly, the established format \mbox{OpenSCENARIO} \cite{Asam2020openSCENARIO} is also designed for virtual closed-loop testing of the planner instead of open-loop testing of the perception subsystem.

\subsection{Generating a Test Scenario Catalog}
\label{sec:scenarios_obtaining}

The ability to describe and structure test scenarios in an ideally formal way is a prerequisite to subsequently generate a representative test scenario catalog. 
Such a test scenario catalog might be part of the specification of the perception subsystem (Sec.~\ref{sec:requirements}). 

The number of test scenarios has to stay feasible, which renders naive sampling of a high-dimensional scenario space unsuitable and instead motivates for the following approaches. 
We distinguish between knowledge-driven and data-driven scenario generation, as already done in~\cite{stellet2015testing, Nalic2020sbt_survey,Neurohr2021criticality}. 
As pointed out by~\cite{Nalic2020sbt_survey}, approaches from both categories usually need to complement each other for a holistic testing strategy.
Likewise, also ISO/PAS 21448 mentions dedicated expert scenario generation (knowledge-driven), as well as large-scale random data recording (data-driven) for the perception verification strategy~\cite[Annex~D]{iso2019sotif}.

In either way, the goal is usually to obtain certain \textit{triggering events}\footnote{or similarly, \textit{triggering conditions}}, which are defined in ISO/PAS 21448~\cite{iso2019sotif} as ``specific conditions of a driving scenario that serve as an initiator for a subsequent system reaction possibly leading to a hazardous event". 
Such a system reaction can occur either in the sensor hardware, in the perception software, or the downstream driving function.
Similar to triggering events, the terms \textit{external influencing factor}~\cite{Bai2019_external_influence_factors_sensing} and \textit{criticality phenomenon}~\cite{Neurohr2021criticality} are also used to describe safety-relevant scenario influences related to perception. 

\subsubsection{Knowledge-Driven Scenario Generation} %
\label{sec:scenario_gen_knowledge_driven}

Since human knowledge of traffic scenarios is often qualitative, knowledge-driven scenario generation usually first yields functional or abstract scenarios. 
Literature on driving scenarios describes the further steps to logical scenarios~\cite{Menzel2019functionalToLogical} and the discretization to concrete scenarios~\cite{Neurohr2020fundamental}.

ISO/PAS 21448 enumerates a list of influencing factors, which shall be used to construct scenarios to identify and evaluate triggering conditions for the perception subsystem \cite[Annex F]{iso2019sotif}. 
This list is included into the standard as an informative example and includes factors like climate, which can be for example fine, cloudy, or rainy.

A specific identification of triggering events according to ISO/PAS 21448 is provided in~\cite{Martin2019} by means of a hazard and operability study (HAZOP). 
The approach is applied to the used sensor modalities camera and lidar as part of an extended automated emergency braking system. 
Sensor modality-specific influencing factors can be analyzed in even more detail by experts using dedicated tests or simulations. 
Example studies analyze the influences of weather, road dirt, or rainfall on lidar (\cite{Rasshofer2011influences},~\cite{Rivero2017roadDirtLidar},~\cite{Berk2019rainfall}, respectively), or corner cases specific to visual perception~\cite{Breitenstein2020visual}.

Such influencing factors or triggering events could also be queried from a domain ontology for driving scenarios~\cite{Bagschik2018_ontology} or from one that also includes perceptual aspects~\cite{Neurohr2021criticality}.
The former source argues that without the use of an ontology, human experts are not likely to generate all possible scenarios based on their knowledge. 
In contrast, if the knowledge is first encoded in an ontology, then the generation of scenarios or scenes from that ontology can cover all possibilities based on the encoded knowledge.

A further approach towards completeness of a scenario catalog is to crowd-source a list of concrete OEDR aspects that are relevant for testing ~\cite{Koopman2019_howMany}. 
The mentioned publication provides a detailed list that is supposed to serve as a starting point for further extension by the community.

While so far, the mentioned influencing factors were selected mostly according to their influence on the perception subsystem, one can also construct perception test scenarios with the downstream driving function in mind. 
This particularly includes scenarios that would be challenging to the driving function in case a perception failure occurs. 
Such a concept is applied under the name of application-oriented test case generation in~\cite{Bai2019_external_influence_factors_sensing} and~\cite{cao2018application} with ACC systems in mind. 
For an ACC function, it appears possible to manually investigate the perception-based influencing factors that negatively affect its behavior. 
In contrast, if the targeted application is a general-purpose urban Level 4 system, it will be harder to enumerate all related influencing factors.

\subsubsection{Data-Driven Scenario Generation}
\label{sec:scenario_gen_data_driven}
If test scenarios are obtained by analyzing large amounts of randomly recorded vehicle fleet data, then the abstraction level of available data is closer to concrete scenarios than to functional scenarios. 
However, the recorded data is initially only available in the data schema of the recording format, which usually does not share the same parameters like a logical scenario space. 
Thus, a challenge in data-driven approaches is to generalize the snippets from the recorded data in a way that they correspond to points in a logical scenario space. 
This is necessary in order to vary individual parameters such that potentially relevant, but not yet observed scenarios can also be generated. 
We are not aware of scientific publications that present data-driven scenario generation in the safety context specifically for perception testing, and thus will refer to related literature about general testing, which has already been reviewed in more detail in~\cite{Nalic2020sbt_survey}. 

One example approach for a data-driven test case generation is described in~\cite{putz2017system}.
In the mentioned publication, a database system of relevant scenarios is described. 
Its input data from various sources such as field operational testing, naturalistic driving studies, accident data, or others, must first be converted to a common input data format of traffic recordings. 
Afterwards, individual snippets of the input data are affiliated to pre-determined logical scenario types, such that each snippet corresponds to one point in a logical scenario space. 
The logical scenario parameters can also be obtained through unsupervised learning on the input data~\cite{Krajewski2018traGAN}. 
If the input data were obtained from large amounts of randomly collected driving data, then one can further compute statistical measures such as the scenario's real-world exposure and its parameters' real-world probability distributions~\cite{putz2017system}.   
By additionally computing criticality metrics on the given snippets, one could estimate the potential severity of a section of the logical scenario space. 
The exposure and severity as meta-information on the scenario space then help to turn the obtained logical scenarios into relevant test cases. 
For example, one could put more testing efforts on scenarios with a high risk, where the risk of a scenario increases with its exposure and severity, and decreases with its controllability~\cite[Appendix A]{Schoener2020challenging}.

\subsubsection{Combined Scenario Generation}
As pointed out by~\cite{Nalic2020sbt_survey}, knowledge-driven and data-driven approaches for scenario generation often complement each other in practice.
For example, the previously mentioned database approach from~\cite{putz2017system} assumes that a concept for logical scenario spaces is given, potentially from expert knowledge.
In turn, knowledge-driven scenario generation can also be backed up by measurements and data~\cite{Nalic2020sbt_survey}.

One example of a combined test case generation for the perception subsystem is given in~\cite{amersbach2017functional}. 
Besides the functional decomposition layers (Fig. ~\ref{fig:amersbach_layers}), the publication also proposes a method for defining particular test cases for each of those decomposition layers. 
The method takes driving scenarios from a database such as~\cite{putz2017system}, assumes an accident if there was none, and then uses a fault tree analysis (FTA) to define pass/fail criteria for the individual subsystems in a given scenario.
This combined approach allows the derivation of perception-specific test scenarios from general driving scenarios. 

\subsubsection{Test Scenarios Specific to the SUT}
In the end, it could also be that the test scenarios that result in the highest risk for a given SUT are \textit{not} found by only considering external influencing factors from the space of all possible scenarios. 
Instead, the unexpected insufficiencies that the DNNs of the SUT have learned might lead to more critical behavior than what a usually suspected external influence could trigger~\cite[Safety Concern~4]{Willers2020safety}. %
For example, \cite{Kurzidem2020analyzingPerceptionArchitectures} identifies perception subsystem inputs that lead to highly uncertain world models for a given SUT by propagating uncertainties through the individual perception components.

\subsubsection{Difficulties in Covering the ODD with Scenarios}

Any attempt to formally describe the ODD and to cover it sufficiently with test scenarios is generally difficult because of the open context problem of the real world~\cite{poddey2019opencontext}. 
Related to this,~\cite[Safety Concern~1]{Willers2020safety} expresses that the data or scenarios used for the development and training of the perception subsystem are usually not a good approximation of the system's ODD in the real open world. 
Furthermore, the real world changes over time, which makes it necessary to iteratively update the ODD description and its coverage with test scenarios~\cite[Safety Concern~2]{Willers2020safety}. 

Moreover, in practice only a finite number of concrete test scenarios can be sampled from the infinite number of concrete scenarios that could be theoretically generated from a logical scenario.
This sampling makes it generally unclear whether the results of two similar concrete test scenarios also hold in between these discrete points. 
In fact, DNNs used for object detection can show a potentially extreme non-linear behavior with respect to only slight input variations for example in weather conditions.
This brittleness, which has been pointed out in~\cite[Safety Concern~6]{Willers2020safety}, makes it difficult to argue about a sufficient ODD coverage with sampled test scenarios.

Nevertheless, combinatorial testing~\cite{Kuhn2009combinatorialTesting}, which is suggested by ISO/PAS 21448 and already applied to computer vision for AVs~\cite{Gladisch2020combinatorial}, can potentially keep the number of discretized concrete tests feasible while providing certain coverage guarantees. 
Furthermore, combinatorial testing that explicitly exploits the functional decomposition approach for AVs~\cite{amersbach2017functional} is performed in~\cite{Weber2020simulation} for simulated AV testing.

\subsection{Executing Scenarios as Test Cases}
\label{sec:executing_scenarios}

We assume the execution of a test scenario to be an offline execution of a perception algorithm on previously recorded raw sensor data. 
In this way, only the sensor mounting positions and sensor hardware are fixed by the recording, whereas multiple perception algorithms can be applied.

For scenarios that correspond to snippets from existing real-world vehicle recordings, the corresponding raw data can directly be used for the test without having to be re-recorded.
Otherwise, the raw sensor data for the test scenarios must be captured. 
Public roads offer less influence on the scenario details, but allow encountering many different scenarios over time with little effort per scenario.
In contrast, proving grounds allow to set up specific scenarios, which however requires high efforts per scenario and may lack realism, e.g., when reproducing precipitation. 

\subsection{Training and Test Sets of Scenarios}
\label{sec:scenarios_different_sets}

If the DNNs of the SUT are trained on data and scenarios that are also used for testing, then overfitting likely takes place during training and an unbiased testing for verification and validation purposes is not possible. 
Thus, it is important to properly separate recorded perception data into individual datasets for training and for testing~\cite[Safety Concern~7]{Willers2020safety},~\cite{SaFAD2019}. 
Furthermore, if the developers evaluate the SUT on a test set multiple times, an unintentional optimization with respect to the test set result might take place over multiple development iterations~\cite{Willers2020safety}.

\subsection{Conclusion of Test Scenarios}
Scenario-based approaches to minimize the overall testing effort of AD functions also seem applicable to the perception subsystem~\cite[Sec.~3.3.4]{SaFAD2019}. 
An agreed-upon description language and data format of perception scenarios would be beneficial for further research.
Various approaches have been proposed to obtain test scenarios, where their combination for a sufficient ODD coverage is an active research topic. 
Furthermore, it is challenging to obtain real-world recordings of all identified test scenarios. 

\section{Reference Data}
\label{sec:axis_ref_data}

The third and final testing axis that this paper covers is the ``knowledge of an ideal result (\textit{reference})"~\cite{stellet2015testing}. 
For object-based environment perception, such an ideal result is an object list containing all road users that the SUT is supposed to perceive. 
We distinguish between the terms \textit{ground truth} and \textit{reference} in a way that a ground truth perfectly and objectively describes the world.
A reference is a subjective approximation of the ground truth in terms of data that are generated by a perception system that is superior to the SUT. 

Requirements for reference systems for environment perception have been analyzed in depth in the context of ADAS~\cite{brahmi2013reference}. 
Apart from the qualitative requirements mobility/portability and reporting of its own uncertainty, the paper states the quantitative requirements reliability, sufficient field-of-view, accuracy, and proper timing of the measurements. 
The following sections review different ways of generating reference data, where each way has its own advantages and disadvantages in the mentioned categories. %
We structure the literature according to the position of the reference sensors, which can be mounted only on the ego vehicle (Sec.~\ref{sec:data_from_ego}), also on other road users (Sec.~\ref{sec:data_from_other_tp}), or externally and not part of any road user (Sec.~\ref{sec:data_from_external}). 
Uncertainty in reference data and an appropriate choice of the reference data source are discussed in Sec.~\ref{sec:ref_data_uncertainty} and Sec.~\ref{sec:ref_data_tradeoffs}, respectively.

\subsection{Reference Data From Ego Vehicle Sensors}
\label{sec:data_from_ego}

The biggest advantage of reference data from sensors of the ego vehicle is that no external measurement equipment is needed. 
Without an external perspective, however, road users that are occluded for the sensors under test can also hardly be present in the reference data. 

\subsubsection{Reference from Sensors Under Test}

Perhaps the most common approach of generating reference data is to use the raw sensor data from the SUT such that humans can label the reference objects and their properties. 
Labels in a sensor-specific coordinate system, such as in a camera's pixel space, need to be transformed to the world model's coordinate system (often in meters and defined over the ground plane) to potentially serve as a reference for it. 
In practice, the human labeling process is often at least semi-automated by offline perception algorithms, which can go back and forth in time.
Such reference data is typically used by datasets for perception algorithm benchmarking such as~\cite{Geiger2012CVPR,caesar2019nuscenes, sun2019waymoopendataset}. 

Using only offline perception algorithms without human intervention (as also done in e.g.~\cite{Robosense2020reference}) dramatically reduces the overall effort and could still result in superior data to the SUT, but generally impairs the data quality.  
A~standardization attempt for human-labeled reference data is ASAM \mbox{OpenLABEL}~\cite{Asam2020openLabel}. 

In either case, no additional sensor hardware is required, which is valid and a big advantage if the perception algorithm is the only component of interest. 
At the same time, it means that the SUT's sensor mounting positions and sensor hardware do not have an independent reference, which might impede their rigorous assessment. 

\subsubsection{Separate Reference Sensors on Ego Vehicle}

Additional sensors on the ego vehicle that are not part of the SUT can potentially also provide a reference for validating the SUT's sensor mounting positions and hardware performance, see e.g.~\cite{Robosense2020reference}. 
This approach seems to be more popular in ADAS developments than for AVs because AVs often already use the best available sensors for their regular operation, and hence there are no other sensors of higher quality left as reference. 

\subsubsection{k-out-of-n-vote of High-Level Fusion Inputs}
\label{sec:ref_data_k_out_of_n}

The present review paper focuses on testing the world model that is generated from fused inputs of all environment sensors. 
In contrast, Berk et al.~\cite{Berk2019ReferenceTruth} propose testing the object-level outputs of individual sensor systems \textit{before} those are fused (more in Sec.~\ref{sec:uncertainty_forecasting}). 
Such testing is needed for the authors' proposed safety assessment strategy for the perception subsystem, where the corresponding reference data could be generated without additional sensors and without manual human effort. 

If for example, a majority of fusion input systems detects a certain road user while one input system does not, then it can be argued that the majority of agreeing input systems provides the reference data for identifying a perception failure in the single disagreeing system. 
This idea is generalized to a so-called k-out-of-n vote of sensor systems. 
It could potentially scale to billions of kilometers of data collection using large vehicle fleets with feasible effort, but its practical implementation on real vehicle data seems to have not yet been described in the available literature. %

\subsection{Reference Data from Other Road Users}
\label{sec:data_from_other_tp}

Other road users can be either equipped with specific test equipment for the estimation of their position and state, or they could communicate their states to the ego vehicle as part of a potential V2X-based regular operation in future. 

\subsubsection{RTK-GNSS-IMUs}
\label{sec:ref_data_rtk_gnss_imu}
Global navigation satellite systems (GNSS) with real-time kinematic correction (RTK) and an attached inertial measurement unit (IMU) have been popular for generating reference data of individual road users. 
Paired systems of those can be installed both in the ego vehicle and at other road users to determine relative positions, velocities, and orientation angles. 
RTK-GNSS can measure the absolute position up to a few centimeters, but face difficulties when the signal is affected by nearby structures such as buildings, trees, or a tunnel. 
IMUs work independently of the surroundings, but only provide relative and incremental positioning information.
Data fusion algorithms can combine the strengths of both components to estimate a consolidated state of the respective road user. 

Despite the size of the measurement hardware, also vulnerable road users (VRUs) can wear such hardware as a backpack~\cite{Scheiner2019gnssVRUs}, which, however, alters their appearance to other sensors. 
In general, only specifically equipped road users can provide reference data using RTK-GNSS-IMUs, which excludes the majority of public road users. %

The accuracy of RTK corrections of GNSS signals in automotive applications has been analyzed in more detail in~\cite{Ho2018rtkGpsAccuracy}. 
Another work has specifically analyzed the accuracy of GNSS systems that are used for relative positioning of local groups of vehicles~\cite{Tahir2018_GPS_Accuracy}. 
For more information on GNSS in general, we refer to the textbook~\cite{ESA2013gnss}.

\subsubsection{Collaborative World Model Through V2X}
\label{sec:collaborative_world_model}
If target vehicles have the capability to accurately localize themselves without dedicated testing hardware, then they could communicate their state to a cloud-based or edge-based collective environment model~\cite{Lampe2019collectiveDriving}. 
The mentioned publication hypothesizes how this could enable a mutual verification of the environment perception of individual road users. 
The key difference to the previously described RTK-GNSS-IMU approaches is that the publication~\cite{Lampe2019collectiveDriving} aims at the regular operation of future production vehicles rather than dedicated testing activities. 

The targeted crowd-sourced reference data generation follows a similar idea like the previously described k-out-of-n-vote of high-level fusion inputs (Sec.~\ref{sec:ref_data_k_out_of_n}). 
The difference here is that the inputs to the vote would originate from vehicles instead of from sensors. 
However, also this approach does not seem to have been demonstrated with real-world vehicles yet. 

\subsection{Reference Data from Non-Road Users}
\label{sec:data_from_external}

The analyzed literature on reference data generation with sensors external to the road users includes stationary infrastructure sensors, unmanned aerial vehicles (UAVs, or drones), and helicopters.
Those data are often intended as either stand-alone naturalistic traffic data or as online enhancements of the AV perception. 
They can still serve as a reference for the SUT if their quality is superior to the SUT according to context-specific reference data requirements. 
A data quality superior to the SUT on the ground can be achieved by a more advantageous bird's eye perspective, by intentional overfitting of the perception algorithms to the sensor's location, and by human error-checking. 
Humans can detect errors in reference object lists easier when comparing them to a raw video from a bird's eye perspective than by comparing them to a potentially disturbed GNSS signal or to potentially incomplete SUT raw data. 

\subsubsection{Reference Data from Stationary Infrastructure Sensors}

Stationary infrastructure sensors are sensors that are mounted on buildings, streetlamps, gantries, bridges, or other nearby structures. 
They are used in urban locations for example in the Ko-PER dataset~\cite{Strigel2014_KoPerDataset}, in the test area autonomous driving Baden-Württemberg~\cite{Fleck2019testAreaBW}, and at the AIM test site~\cite{knake2016aimintersection}.
Infrastructure sensors at highways or motorways are used in~\cite{notz2020extraction} and at the Providentia sensor system~\cite{Kraemmer2019providentia}, where the latter source assesses its performance using a helicopter-based system, which is covered in Sec.~\ref{sec:ref_data_helicopters}. 
The projects HDVMess and ACCorD aim at installing mobile and fixed infrastructure sensors at various sites~\cite{Kloeker2020infrastructure_ACK} and evaluate their data generation with a UAV-based reference system~\cite{Kloeker2020trafficRecording}. 

\subsubsection{Reference Data from UAVs}
\label{sec:ref_data_UAVs}

Camera-based reference data generation from a bird's eye perspective has been investigated at least since 2014~\cite{Kubertschak2014kamerabasiertes}. 
The necessary processing steps to track ground vehicles from UAVs were later described in more detail in~\cite{Guido2016uav} and~\cite{Kruber2020uav}. 
UAVs are used as reference sensors for sensor modeling in~\cite{Krajewski2020UsingDrones} and for assessing infrastructure sensors in~\cite{Kloeker2020trafficRecording}. 
A qualitative assessment of UAV-based reference systems in contrast to other reference systems is given in~\cite{Krajewski2019dronesfortesting}. 
Reference data from a UAV's perspective tend to suffer from less occlusions than from perspectives closer to the ground, however, new sorts of occlusions from treetops or bridges occur. 

\subsubsection{Reference Data from Helicopters}
\label{sec:ref_data_helicopters}

Additionally, helicopters have been used to record the ground traffic for verification and validation purposes of AD systems. 
The DLRAD dataset~\cite{kurz2018dlrad} was partially recorded using a helicopter that follows a target vehicle with active environment perception. 
A previous publication from the same project~\cite{Kurz2015helicopter} describes how such a helicopter-based system can be used to validate the behavior of ADAS. 
Furthermore, a helicopter-based reference data generation is also described and used in~\cite{Kraemmer2019providentia} to assess a static infrastructure sensor system.

\subsection{Uncertainty in Reference Data}
\label{sec:ref_data_uncertainty}
Reference systems do not provide an absolute ground truth, but only have to come closer to it than the SUT does. 
This has been pointed out and quantified with respect to labeling uncertainty in~\cite{wang2020inferring}. %
Thus, the quality of the reference data (or labels) should be taken into account such that test results are not misleading~\cite[Safety Concern~8]{Willers2020safety}. 

For automatically generated reference data from external sensors, there might not be labeling uncertainty, but uncertainty in temporal synchronization of reference and SUT data, as well as uncertainty in their spatial alignment. %

Moreover, any automatically generated reference data without human error-checking can suffer from unexpected inaccuracies in the reference measurement system. 
Those inaccuracies might be rare enough to not matter in small-scale R\&D testing, but over large amounts of test scenarios or kilometers, as needed for safety assurance, the probability of encountering relevant reference system errors increases. 

\subsection{Choice of Reference Data Source}
\label{sec:ref_data_tradeoffs}

Different ways of generating reference data are used for different aspects of perception testing, where each way has its unique advantages. 
For example, human labels might be best for testing what a perception algorithm can infer from given raw data in the absence of additional data sources. 
Separate high-quality sensors on the ego vehicle could be the most useful choice for benchmarking individually inferior sensor systems. 
RTK-GNSS-IMUs might be best for testing the perception of small numbers of vehicles in the absence of high buildings. 
Data from an aerial perspective might be most suitable for testing the perception performance under static and dynamic occlusions in complex naturalistic traffic. 
Infrastructure test fields that spread over a large area can be best suited for investigations that require a large field of view.
Finally, reference data from the \mbox{k-out-of-n-vote} might have the best potential for keeping the testing effort feasible for large amounts of kilometers or scenarios. 

\subsection{Conclusion of Reference Data}
\label{sec:ref_data_conclusion}

The most appropriate choice of reference data depends on the detailed objectives of the respective testing activity within the safety case (\cite[Sec.~3.6.3]{SaFAD2019}). 
In any case, the reference data's limitations should be considered. 

\section{Discussion of Review Results}
\label{sec:discussion}

Sec.~\ref{sec:discussion_answers} summarizes the particular conclusions of each testing axis in terms of the largest issues regarding the primary research question. %
Intersection topics between the axes are covered in Sec.~\ref{sec:discussion_interdependencies}. %
Finally, testing as reviewed in this paper is likely necessary, but not sufficient for safety assurance of AVs.  
Therefore, Sec.~\ref{sec:discussion_other_activities} places the reviewed topics into a broader context by drawing connections to related activities around perceptual aspects in safety assurance. 

\subsection{Open Issues per Testing Axis}
\label{sec:discussion_answers}

In terms of metrics and test criteria (Sec.~\ref{sec:axis_criteria_metrics}), current challenges are the incompleteness of criteria due to the open-world context and also the consideration of a potential safety impact in the metrics. %

Open issues about perception-specific test scenarios (Sec.~\ref{sec:axis_test_scenarios}) 
are a common description language and format, a combination of scenario generation approaches for sufficient ODD coverage, and the difficulty of recording the identified test scenarios in the real world.

The most appropriate source of reference data (Sec.~\ref{sec:axis_ref_data}) depends on the investigated aspect of environment perception. 
A combination of multiple such investigations into a harmonized methodology targeting the safety proof appears to be not yet demonstrated in the literature. 

The mentioned open issues from the literature might be the reason why as of now, safety standards (Sec.~\ref{sec:standards}) leave many details around perception testing open. 

\subsection{Open Issues Between the Testing Axes}
\label{sec:discussion_interdependencies}

Solutions to the individual testing axes must be able to fit into a consolidated methodology that also considers intersection topics between the axes.
For example, at the intersection of test criteria and scenarios, the criteria on the SUT's FOV might be relaxed if the ego vehicle drives slowly due to rain in the respective scenario. 
In between the axes scenarios and reference data, the fact that rain is present in the scenario might affect some ways of reference data generation more than others. 
The intersection between reference data and metrics/criteria includes the metrics' sensitivity to details in the reference data.
For example, a metric that penalizes false negatives behind occlusions produces different results depending on whether the reference data contain these occluded objects or not. 

While an all-encompassing methodology that covers all of such details seems to be not yet demonstrated, approaches such as~\cite{Czarnecki_2018_Framework} or~\cite{Berk2020summary} already propose general ways of how the dependability of the perception subsystem could be managed and tested. 

\subsection{Further Safety Assurance Activities Regarding Perception} %
\label{sec:discussion_other_activities}
Testing as described in this review (see Table ~\ref{table:focus_of_paper}) is only one activity out of many within an all-encompassing safety argumentation. 
The following subsections intend to reveal this larger context by drawing connections to related activities in perception dependability.
We structure the approaches according to the means for attaining dependability by \cite{Avizienis2004dependability}, which have recently been adopted to uncertainties in AD \cite{Gansch2020uncertainty}. 
They consist of prevention, removal, tolerance, and forecasting of faults or uncertainty, respectively. 

\subsubsection{Uncertainty Prevention}
\label{sec:uncertainty_prevention}

\textit{Formal verification} of the perception subsystem could ideally prevent the occurrence of certain world model uncertainties.
So far, we are not aware of formal verification methods that cover the entire perception subsystem. 
Also, the review paper~\cite{Riedmaier2020_surveySBT} only mentions formal verification methods for the downstream driving function. %
However, formal analysis and verification of DNNs for perception tasks is an active topic of research~\cite{Wang2018formal,Seshia2016verifiedAI}.
Moreover, the way in which uncertain world models affect formally verified downstream driving functions is actively being researched, for example in the context of RSS~\cite{Salay2020_purss}. 

Another means of uncertainty prevention is the restriction to certain ODDs~\cite{Gansch2020uncertainty}.

\subsubsection{Uncertainty Removal}
\label{sec:uncertainty_removal}

Perceptual uncertainties can be removed during development and during its use after the initial release. 
Uncertainty removal during development includes safety analysis methods such as fault tree analysis (FTA)~\cite{Gansch2020uncertainty}, or insightful developer testing with the goal of reducing found uncertainties afterwards. 
For example, \textit{white-box-testing of the perception subsystem} can reveal uncertainties related to the physically complex measurement principles (see e.g.~\cite{holder2019modeling}). 
Human-understandable insights into the DNNs for object detection are, however, difficult~\cite[Safety Concern~3]{Willers2020safety}.

\textit{Testing on simulated raw data} %
allows the perception algorithm to be tested without the need for physical sensor hardware
(e.g.~\cite{Barbier2019PerceptionValidation} for grid mapping). 
For testing object perception algorithms, a simulation environment such as \cite{NeumannCosel2014vtd} could produce both the raw sensor data as an input to the algorithm under test and a ground truth object list as a reference.
A hybrid approach is to test the perception algorithm on recorded real data with simulated injected faults~\cite{Rao2019faultInjection}. 

After the initial release, \textit{incremental safety assurance} through system updates is a necessary means to remove uncertainties because the ODD of the AV changes continuously~\cite[Safety Concern~2]{Willers2020safety}. 
Ideally, the safety argumentation should allow minor system updates without having to repeat major testing efforts~\cite{Koopman2018towardFramework}. 
Agile safety cases~\cite{Myklebust2020agile} could enable the safe deployment of such incremental updates through a DevOps-like workflow~\cite{Czarnecki2019devOps}.

\subsubsection{Uncertainty Tolerance}
\label{sec:uncertainty_tolerance}

\textit{Runtime monitoring} of the uncertainty of the produced world model can help mitigating the safety consequences of perception errors~\cite{Antonante2020perceptionMonitoringv3}. 
An RSS-compliant planner would become more cautious for uncertain world models due to a reduced set of guaranteed safe control actions~\cite{Salay2020_purss} (Sec.~\ref{sec:downstream_comparison}).
Similarly, in a service-oriented architecture, sensor data that is monitored as too uncertain or faulty could lead to a bypass of the usual information processing pipeline in favor of a safe behavior for degraded modes~\cite[Fig.~3]{Woopen2018UNICARagil}.
An overview of self-representation and monitoring of subsystems is given in~\cite{Nolte2020selfRepresentation}. %
Similarly, in the ISO 26262 context, it was proposed that the perception subsystem dynamically outputs one world model for each ASIL level such that the planner can dynamically take just that world model that it needs to comply to the ASIL requirement of the current driving situation~\cite{Johansson2016perceptionASIL,Johansson2017assessingUseOfSensors, Cassel2020SAEPerceptionRequirements}.
In any of such applications, it is crucial to assure that the reported uncertainties are correctly calibrated (Sec.~\ref{sec:self_reporting_metrics}).

Furthermore, the mentioned \textit{collaborative world model through V2X} (Sec.~\ref{sec:collaborative_world_model}) can also serve as a means for uncertainty tolerance if participating cars can reduce their world model uncertainties through, e.g., a cloud-based fusion~\cite{Lampe2020reducing}.

\subsubsection{Uncertainty Forecasting}
\label{sec:uncertainty_forecasting}

\begin{figure}[t]
    \centering
    \vspace*{2mm}
    \includegraphics[width=\linewidth]{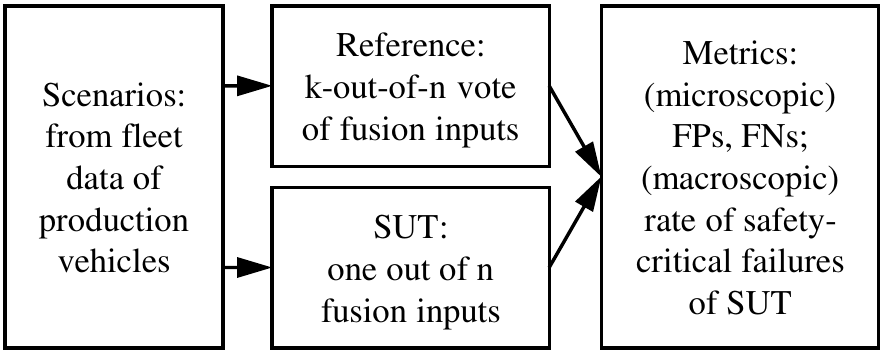}
    \caption{Perception testing for macroscopic safety assurance without separately generated reference data~\cite{Berk2020summary,Berk2019ReferenceTruth}, expressed in taxonomy (Fig.~\ref{fig:taxonomy_general}).}
    \label{fig:taxonomy_berk}
\end{figure}

The failure rates in a fused world model must be extremely low, which means that the direct test effort for proving or forecasting it is extremely high.

One approach to overcome this so-called approval trap \cite{Winner2016quoVadis} is to test the more failure-prone world models of individual sensor systems \textit{before} they are fused~\cite{shalevshwartz2017formalRSS},~\cite{Berk2020summary}.
The latter source investigates how this could be realized without separately generated reference data, allowing the tests to be potentially executed in shadow-mode by production vehicles. 
In this approach (Fig.~\ref{fig:taxonomy_berk}), the reference data take the form of a \mbox{k-out-of-n}-vote of object-level fusion inputs (Sec.~\ref{sec:ref_data_k_out_of_n}). 
Such a mutual cross-referencing mechanism is also suggested in \textit{Safety First for Automated Driving} \cite[Sec.~2.2.2.4]{SaFAD2019}. 

The mentioned testing approach exploits the redundancy of fusion inputs in the following way~\cite{berk2019exploiting}. 
If a mean time between failures of for example $\SI{e4}{\hour}$ can be demonstrated for two statistically independent inputs to a data fusion system, then the fusion output has a mean time between failures of about $\SI{e8}{\hour}$. 
Such a redundancy exploitation has also been stated in the RSS papers~\cite{shalevshwartz2017formalRSS},~\cite{Weast2020} and in literature on perception subsystem architectures~\cite{Grubmuller2019robustFusion},~\cite{kaprocki2019multiunit} and perception requirements~\cite{Cassel2020SAEPerceptionRequirements}. 
The actual number of necessary test kilometers is then computed in a statistically sound way based on how statistically dependent the inputs to the object-level fusion system actually are~\cite{Berk2017BayesianTestDesign, Bock2016testEffortEstimation}.

Whether or not perceptual uncertainties lead to safety-critical failures on the vehicle level might require a \textit{sensitivity analysis of the driving function}.
Reachability analysis (e.g.~\cite{althoff2010reachability, Pek2020onlineVerification}) could be used to analyze in detail whether an uncertain world model can lead to unsafe future states, as outlined in the context of the \textit{incurred severity} metric~\cite{Salay2019_safety_perceptual_components}.
An impact analysis of perceptual errors on the downstream driving function has also already been described in~\cite{Piazzoni2020Modeling_conference}.

Furthermore, uncertainty forecasting for the release argumentation could make use of the tests mentioned under \textit{Uncertainty Removal} (Sec.~\ref{sec:uncertainty_removal}) if they are validated and executed on separate test datasets. 

\section{Conclusion}
\label{sec:conclusion}
We analyzed literature from multiple neighboring fields, attempting to provide a structured overview of safety-relevant testing activities for the environment perception of AVs.
These neighboring fields included, but are not limited to, AV safety assurance, perception algorithm development, and safeguarding of artificial intelligence. 
A combined literature search process that consisted of undocumented search, keyword-based search and snowballing search required large efforts, but appeared to be suitable for covering the relevant literature. 

The analyzed test methods from the available literature seem to be not yet capable of demonstrating the dependability of the perception subsystem for series-production Level~4 vehicles. %
To overcome this issue, the provided overview can serve as a common basis for a further harmonization of primary literature contributions into all-encompassing novel testing methodologies.

\section*{Acknowledgement}
\label{sec:contributions}

The authors would like to thank their colleagues at the Institute for Automotive Engineering and their project partners of the project VVM, a project of the PEGASUS family, for their valuable feedback on the initial version of this paper.
Furthermore, the authors would like to thank Robert Krajewski for fruitful discussions.

Despite all efforts, we might have unintentionally overlooked individual literature contributions to the research question. In that case, we apologize to the corresponding authors and are welcoming such feedback from the community.

\bibliographystyle{IEEEtran} %
\bibliography{IEEEabrv,AllSourcesMHO,allSourcesMSO}

\begin{thebibliography}{100}
\providecommand{\url}[1]{#1}
\csname url@rmstyle\endcsname
\providecommand{\newblock}{\relax}
\providecommand{\bibinfo}[2]{#2}
\providecommand\BIBentrySTDinterwordspacing{\spaceskip=0pt\relax}
\providecommand\BIBentryALTinterwordstretchfactor{4}
\providecommand\BIBentryALTinterwordspacing{\spaceskip=\fontdimen2\font plus
\BIBentryALTinterwordstretchfactor\fontdimen3\font minus
  \fontdimen4\font\relax}
\providecommand\BIBforeignlanguage[2]{{%
\expandafter\ifx\csname l@#1\endcsname\relax
\typeout{** WARNING: IEEEtran.bst: No hyphenation pattern has been}%
\typeout{** loaded for the language `#1'. Using the pattern for}%
\typeout{** the default language instead.}%
\else
\language=\csname l@#1\endcsname
\fi
#2}}

\bibitem{SAE18}
SAE, ``J3016: Taxonomy and definitions for terms related to driving automation
  systems for on-road motor vehicles,'' 2018.

\bibitem{Dietmayer2014representation}
\BIBentryALTinterwordspacing
K.~C.~J. Dietmayer, S.~Reuter, and D.~Nuss, \emph{Representation of Fused
  Environment Data}.\hskip 1em plus 0.5em minus 0.4em\relax Cham: Springer
  International Publishing, 2014, pp. 1--30. [Online]. Available:
  \url{https://doi.org/10.1007/978-3-319-09840-1_25-1}
\BIBentrySTDinterwordspacing

\bibitem{dietmayer2016predicting}
K.~Dietmayer, ``Predicting of machine perception for automated driving,'' in
  \emph{Autonomous Driving}.\hskip 1em plus 0.5em minus 0.4em\relax Springer,
  2016, pp. 407--424.

\bibitem{stellet2015testing}
J.~E. Stellet, M.~R. Zofka, J.~Schumacher, T.~Schamm, F.~Niewels, and J.~M.
  Z{\"o}llner, ``Testing of advanced driver assistance towards automated
  driving: A survey and taxonomy on existing approaches and open questions,''
  in \emph{IEEE 18th Int. Conf. on Intelligent Transportation Systems}.\hskip
  1em plus 0.5em minus 0.4em\relax IEEE, 2015, pp. 1455--1462.

\bibitem{amersbach2017functional}
C.~Amersbach and H.~Winner, ``Functional decomposition: An approach to reduce
  the approval effort for highly automated driving,'' in \emph{8. Tagung
  Fahrerassistenz}.\hskip 1em plus 0.5em minus 0.4em\relax Munich: Technical
  University of Munich, 2017.

\bibitem{Koopman2018towardFramework}
P.~Koopman and M.~Wagner, ``Toward a framework for highly automated vehicle
  safety validation,'' in \emph{{SAE} Technical Paper Series}.\hskip 1em plus
  0.5em minus 0.4em\relax {SAE} International, Apr. 2018.

\bibitem{iso2019sotif}
\BIBentryALTinterwordspacing
\emph{ISO/PAS 21448:2019 Road vehicles — Safety of the intended
  functionality}, ISO Std., Jan. 2019. [Online]. Available:
  \url{https://www.iso.org/standard/70939.html}
\BIBentrySTDinterwordspacing

\bibitem{VVM2020Begriffsregister}
``Glossary of research project `{VVM - Verification and Validation Methods for
  Automated Vehicles Level 4 and 5}','' Internal, Nov. 2020.

\bibitem{leveson2016engineering}
N.~G. Leveson, \emph{Engineering a safer world: Systems thinking applied to
  safety}.\hskip 1em plus 0.5em minus 0.4em\relax The MIT Press, 2016.

\bibitem{ISO_26262_2018}
\emph{ISO 26262 Road vehicles - Functional safety}, ISO Std., 2018.

\bibitem{JCGM2008GUM}
\emph{Evaluation of measurement data — Guide to the expression ofuncertainty
  in measurement}, Joint Committee for Guides in Metrology Std., Rev. GUM 1995
  with minor corrections, 2008.

\bibitem{Ulbrich2015}
S.~Ulbrich, T.~Menzel, A.~Reschka, F.~Schuldt, and M.~Maurer, ``Defining and
  substantiating the terms scene, situation, and scenario for automated
  driving,'' in \emph{{IEEE} 18th Int. Conf. on Intelligent Transportation
  Systems}.\hskip 1em plus 0.5em minus 0.4em\relax {IEEE}, Sept. 2015.

\bibitem{Junietz2019DissRiskMetrics}
\BIBentryALTinterwordspacing
P.~M. Junietz, ``Microscopic and macroscopic risk metrics for the safety
  validation of automated driving,'' Ph.D. dissertation, TU Darmstadt,
  Darmstadt, 2019. [Online]. Available:
  \url{http://tuprints.ulb.tu-darmstadt.de/9282/}
\BIBentrySTDinterwordspacing

\bibitem{Stellet2020ValidationSurvey}
J.~E. Stellet, M.~Woehrle, T.~Brade, A.~Poddey, and W.~Branz, ``Validation of
  automated driving--a structured analysis and survey of approaches,'' in
  \emph{13. Uni-DAS e.V. Workshop Fahrerassistenz und automatisiertes Fahren},
  2020.

\bibitem{Riedmaier2020_surveySBT}
S.~Riedmaier, T.~Ponn, D.~Ludwig, B.~Schick, and F.~Diermeyer, ``Survey on
  scenario-based safety assessment of automated vehicles,'' \emph{{IEEE}
  Access}, vol.~8, pp. 87\,456--87\,477, 2020.

\bibitem{Nalic2020sbt_survey}
D.~Nalic, T.~Mihalj, M.~Baeumler, M.~Lehmann, A.~Eichberger, and
  S.~Bernsteiner, ``Scenario based testing of automated driving systems: A
  literature survey,'' in \emph{FISITA Web Contress}, Oct. 2020.

\bibitem{Pegasus2019Overview}
\BIBentryALTinterwordspacing
``Pegasus method: An overview,'' Online, 2019. [Online]. Available:
  \url{https://www.pegasusprojekt.de/files/tmpl/Pegasus-Abschlussveranstaltung/PEGASUS-Gesamtmethode.pdf}
\BIBentrySTDinterwordspacing

\bibitem{borg2018safely}
M.~Borg, C.~Englund, K.~Wnuk, B.~Duran, C.~Levandowski, S.~Gao, Y.~Tan,
  H.~Kaijser, H.~L{\"o}nn, and J.~T{\"o}rnqvist, ``Safely entering the deep: A
  review of verification and validation for machine learning and a challenge
  elicitation in the automotive industry,'' \emph{arXiv:1812.05389}, 2018.

\bibitem{marti2019review}
E.~Marti, M.~A. de~Miguel, F.~Garcia, and J.~Perez, ``A review of sensor
  technologies for perception in automated driving,'' \emph{IEEE Intelligent
  Transportation Systems Magazine}, vol.~11, no.~4, pp. 94--108, 2019.

\bibitem{Rosique2019}
F.~Rosique, P.~J. Navarro, C.~Fern{\'{a}}ndez, and A.~Padilla, ``A systematic
  review of perception system and simulators for autonomous vehicles
  research,'' \emph{Sensors}, vol.~19, no.~3, p. 648, Feb. 2019.

\bibitem{Zhu2017PerceptionOverview}
H.~Zhu, K.-V. Yuen, L.~Mihaylova, and H.~Leung, ``Overview of environment
  perception for intelligent vehicles,'' \emph{{IEEE} Transactions on
  Intelligent Transportation Systems}, vol.~18, no.~10, pp. 2584--2601, Oct.
  2017.

\bibitem{Willers2020safety}
O.~Willers, S.~Sudholt, S.~Raafatnia, and S.~Abrecht, ``Safety concerns and
  mitigation approaches regarding the use of deep learning in safety-critical
  perception tasks,'' in \emph{Computer Safety, Reliability, and Security.
  SAFECOMP Workshops}, A.~Casimiro, F.~Ortmeier, E.~Schoitsch, F.~Bitsch, and
  P.~Ferreira, Eds.\hskip 1em plus 0.5em minus 0.4em\relax Cham: Springer
  International Publishing, 2020, pp. 336--350.

\bibitem{Burton2017SafetyML_HAD}
S.~Burton, L.~Gauerhof, and C.~Heinzemann, ``Making the case for safety of
  machine learning in highly automated driving,'' in \emph{Computer Safety,
  Reliability, and Security}, S.~Tonetta, E.~Schoitsch, and F.~Bitsch,
  Eds.\hskip 1em plus 0.5em minus 0.4em\relax Cham: Springer International
  Publishing, 2017, pp. 5--16.

\bibitem{Kitchenham2007guidelinesSLR}
B.~Kitchenham and S.~Charters, ``Guidelines for performing systematic
  literature reviews in software engineering,'' Evidence-Based Software
  Engineering (EBSE) Project, Tech. Rep. v2.3, 2007.

\bibitem{wohlin2014snowballing}
C.~Wohlin, ``Guidelines for snowballing in systematic literature studies and a
  replication in software engineering,'' in \emph{Proc. of the 18th Int. Conf.
  on Evaluation and Assessment in Software Engineering}, 2014, pp. 1--10.

\bibitem{Schalling2019LidarBenchmarking}
F.~{Schalling}, S.~{Ljungberg}, and N.~{Mohan}, ``Benchmarking lidar sensors
  for development and evaluation of automotive perception,'' in \emph{4th Int.
  Conf. and Workshops on Recent Advances and Innovations in Engineering
  (ICRAIE)}, 2019, pp. 1--6.

\bibitem{Salay2020_purss}
R.~Salay, K.~Czarnecki, I.~Alvarez, M.~S. Elli, S.~Sedwards, and J.~Weast,
  ``{PURSS}: Towards perceptual uncertainty aware responsibility sensitive
  safety with {ML},'' in \emph{Proc. of the Workshop on Artificial Intelligence
  Safety (SafeAI)}, New York, USA, 2020.

\bibitem{Salay2019_safety_perceptual_components}
R.~Salay, M.~Angus, and K.~Czarnecki, ``A safety analysis method for perceptual
  components in automated driving,'' \emph{IEEE 30th Int. Symposium on Software
  Reliability Engineering (ISSRE)}, pp. 24--34, 2019.

\bibitem{Salay2019partialspecifications}
R.~Salay and K.~Czarnecki, ``Improving {ML} safety with partial
  specifications,'' in \emph{Computer Safety, Reliability, and Security},
  A.~Romanovsky, E.~Troubitsyna, I.~Gashi, E.~Schoitsch, and F.~Bitsch,
  Eds.\hskip 1em plus 0.5em minus 0.4em\relax Cham: Springer International
  Publishing, 2019, pp. 288--300.

\bibitem{Czarnecki_2018_Framework}
\BIBentryALTinterwordspacing
K.~Czarnecki and R.~Salay, ``Towards a framework to manage perceptual
  uncertainty for safe automated driving,'' \emph{Lecture Notes in Computer
  Science}, p. 439–445, 2018. [Online]. Available:
  \url{http://dx.doi.org/10.1007/978-3-319-99229-7_37}
\BIBentrySTDinterwordspacing

\bibitem{berk2019exploiting}
M.~Berk, O.~Schubert, H.-M. Kroll, B.~Buschardt, and D.~Straub, ``Exploiting
  redundancy for reliability analysis of sensor perception in automated driving
  vehicles,'' \emph{IEEE Transactions on Intelligent Transportation Systems},
  2019.

\bibitem{Berk2019ReferenceTruth}
------, ``Reliability assessment of safety-critical sensor information: Does
  one need a reference truth?'' \emph{{IEEE} Transactions on Reliability},
  vol.~68, no.~4, pp. 1227--1241, Dec. 2019.

\bibitem{shalevshwartz2017formalRSS}
S.~Shalev-Shwartz, S.~Shammah, and A.~Shashua, ``On a formal model of safe and
  scalable self-driving cars,'' \emph{arXiv:1708.06374}, 2017, v6.

\bibitem{brahmi2013reference}
M.~Brahmi, K.-H. Siedersberger, A.~Siegel, and M.~Maurer, ``Reference systems
  for environmental perception: Requirements, validation and metric-based
  evaluation,'' in \emph{6. Tagung Fahrerassistenzsysteme}, 2013.

\bibitem{Cassel2020SAEPerceptionRequirements}
\BIBentryALTinterwordspacing
A.~Cassel, C.~Bergenhem, O.~M. Christensen, H.-M. Heyn, S.~Leadersson-Olsson,
  M.~Majdandzic, P.~Sun, A.~Thorsén, and J.~Trygvesson, ``On perception safety
  requirements and multi sensor systems for automated driving systems,'' in
  \emph{SAE Technical Paper}.\hskip 1em plus 0.5em minus 0.4em\relax SAE
  International, Apr. 2020. [Online]. Available:
  \url{https://doi.org/10.4271/2020-01-0101}
\BIBentrySTDinterwordspacing

\bibitem{Berk2019Dissertation}
M.~Berk, ``Safety assessment of environment perception in automated driving
  vehicles,'' Dissertation, Technical University Munich, Munich, 2019.

\bibitem{Salay2017MLIso26262}
R.~Salay, R.~Queiroz, and K.~Czarnecki, ``An analysis of {ISO} 26262: Using
  machine learning safely in automotive software,'' \emph{arXiv:1709.02435},
  2017.

\bibitem{tas2016functional}
{\"O}.~{\c{S}}. Ta{\c{s}}, F.~Kuhnt, J.~M. Z{\"o}llner, and C.~Stiller,
  ``Functional system architectures towards fully automated driving,'' in
  \emph{IEEE Intelligent Vehicles Symposium (IV)}.\hskip 1em plus 0.5em minus
  0.4em\relax IEEE, 2016, pp. 304--309.

\bibitem{kaprocki2019multiunit}
N.~Kaprocki, G.~Velikic, N.~Teslic, and M.~Krunic, ``Multiunit automotive
  perception framework: Synergy between {AI} and deterministic processing,'' in
  \emph{{IEEE} 9th Int. Conf. on Consumer Electronics ({ICCE}-Berlin)}.\hskip
  1em plus 0.5em minus 0.4em\relax {IEEE}, Sept. 2019.

\bibitem{Martin2019}
H.~{Martin}, B.~{Winkler}, S.~{Grubmüller}, and D.~{Watzenig},
  ``Identification of performance limitations of sensing technologies for
  automated driving,'' in \emph{IEEE Int. Conf. on Connected Vehicles and Expo
  (ICCVE)}, 2019, pp. 1--6.

\bibitem{Bai2019_external_influence_factors_sensing}
J.~{Bai}, Y.~{Zhan}, P.~{Cao}, L.~{Huang}, Y.~{Xu}, and X.~{Bi}, ``Application
  oriented identification of external influence factors on sensing for
  validation of automated driving systems,'' in \emph{IEEE MTT-S International
  Wireless Symposium (IWS)}, 2019, pp. 1--3.

\bibitem{ulbrich2017functional}
S.~Ulbrich, A.~Reschka, J.~Rieken, S.~Ernst, G.~Bagschik, F.~Dierkes, M.~Nolte,
  and M.~Maurer, ``Towards a functional system architecture for automated
  vehicles,'' \emph{arXiv:1703.08557}, 2017.

\bibitem{althoff2010reachability}
M.~Althoff, ``Reachability analysis and its application to the safety
  assessment of autonomous cars,'' Ph.D. dissertation, Technical University
  Munich, 2010.

\bibitem{Bagschik2018_ontology}
G.~Bagschik, T.~Menzel, and M.~Maurer, ``Ontology based scene creation for the
  development of automated vehicles,'' in \emph{{IEEE} Intelligent Vehicles
  Symposium ({IV})}.\hskip 1em plus 0.5em minus 0.4em\relax {IEEE}, June 2018.

\bibitem{Burton2019confidence_arguments}
S.~Burton, L.~Gauerhof, B.~B. Sethy, I.~Habli, and R.~Hawkins, ``Confidence
  arguments for evidence of performance in machine learning for highly
  automated driving functions,'' in \emph{Computer Safety, Reliability, and
  Security}, A.~Romanovsky, E.~Troubitsyna, I.~Gashi, E.~Schoitsch, and
  F.~Bitsch, Eds.\hskip 1em plus 0.5em minus 0.4em\relax Cham: Springer
  International Publishing, 2019, pp. 365--377.

\bibitem{Koopman2019_howMany}
P.~Koopman and F.~Fratrik, ``How many operational design domains, objects, and
  events?'' in \emph{SafeAI@AAAI}, 2019.

\bibitem{Kubertschak2014kamerabasiertes}
T.~Kubertschak, M.~Wittenzellner, and M.~Maehlisch, ``{Kamerabasiertes
  Referenzsystem für Fahrerassistenzsysteme},'' in \emph{Forum
  Bildverarbeitung}, vol.~82.\hskip 1em plus 0.5em minus 0.4em\relax KIT
  Scientific Publishing, 2014, p. 261.

\bibitem{Strigel2014_KoPerDataset}
E.~Strigel, D.~Meissner, F.~Seeliger, B.~Wilking, and K.~Dietmayer, ``The
  {Ko-PER} intersection laserscanner and video dataset,'' in \emph{17th
  International {IEEE} Conf. on Intelligent Transportation Systems
  ({ITSC})}.\hskip 1em plus 0.5em minus 0.4em\relax {IEEE}, Oct. 2014.

\bibitem{Berk2020summary}
M.~Berk, O.~Schubert, H.-M. Kroll, B.~Buschardt, and D.~Straub, ``Assessing the
  safety of environment perception in automated driving vehicles,'' \emph{{SAE}
  International Journal of Transportation Safety}, vol.~8, no.~1, Apr. 2020.

\bibitem{Berk2017BayesianTestDesign}
M.~Berk, H.-M. Kroll, O.~Schubert, B.~Buschardt, and D.~Straub, ``Bayesian test
  design for reliability assessments of safety-relevant environment sensors
  considering dependent failures,'' in \emph{{SAE} Technical Paper
  Series}.\hskip 1em plus 0.5em minus 0.4em\relax {SAE} International, Mar.
  2017.

\bibitem{Berk2019rainfall}
M.~Berk, M.~Dura, J.~V. Rivero, O.~Schubert, H.-M. Kroll, B.~Buschardt, and
  D.~Straub, ``A stochastic physical simulation framework to quantify the
  effect of rainfall on automotive lidar,'' in \emph{{SAE} Technical Paper
  Series}.\hskip 1em plus 0.5em minus 0.4em\relax {SAE} International, Apr.
  2019.

\bibitem{Johansson2016perceptionASIL}
R.~Johansson and J.~Nilsson, ``The need for an environment perception block to
  address all {ASIL} levels simultaneously,'' in \emph{{IEEE} Intelligent
  Vehicles Symposium ({IV})}.\hskip 1em plus 0.5em minus 0.4em\relax {IEEE},
  June 2016.

\bibitem{Johansson2017assessingUseOfSensors}
R.~Johansson, S.~Alissa, S.~Bengtsson, C.~Bergenhem, O.~Bridal, A.~Cassel,
  D.-J. Chen, M.~Gassilewski, J.~Nilsson, A.~Sandberg, S.~Ursing, F.~Warg, and
  A.~Werneman, ``A strategy for assessing safe use of sensors in autonomous
  road vehicles,'' in \emph{Computer Safety, Reliability, and Security},
  S.~Tonetta, E.~Schoitsch, and F.~Bitsch, Eds.\hskip 1em plus 0.5em minus
  0.4em\relax Cham: Springer International Publishing, 2017, pp. 149--161.

\bibitem{Bock2016testEffortEstimation}
F.~Bock, S.~Siegl, and R.~German, ``Mathematical test effort estimation for
  dependability assessment of sensor-based driver assistance systems,'' in
  \emph{42th Euromicro Conf. on Software Engineering and Advanced Applications
  ({SEAA})}.\hskip 1em plus 0.5em minus 0.4em\relax {IEEE}, Aug. 2016.

\bibitem{Rivero2017roadDirtLidar}
J.~R.~V. Rivero, I.~Tahiraj, O.~Schubert, C.~Glassl, B.~Buschardt, M.~Berk, and
  J.~Chen, ``Characterization and simulation of the effect of road dirt on the
  performance of a laser scanner,'' in \emph{{IEEE} 20th Int. Conf. on
  Intelligent Transportation Systems ({ITSC})}.\hskip 1em plus 0.5em minus
  0.4em\relax {IEEE}, Oct. 2017.

\bibitem{Rasshofer2011influences}
R.~H. Rasshofer, M.~Spies, and H.~Spies, ``Influences of weather phenomena on
  automotive laser radar systems,'' \emph{Advances in Radio Science: ARS},
  vol.~9, pp. 49--60, 2011.

\bibitem{Fawcett2006ROC}
T.~Fawcett, ``An introduction to {ROC} analysis,'' \emph{Pattern Recognition
  Letters}, vol.~27, no.~8, pp. 861--874, June 2006.

\bibitem{cao2018application}
P.~Cao and L.~Huang, ``Application oriented testcase generation for validation
  of environment perception sensor in automated driving systems,'' \emph{SAE
  MOBILUS}, 2018.

\bibitem{Czarnecki2018OntologyPart2}
K.~Czarnecki, ``\BIBforeignlanguage{en}{Operational world model ontology for
  automated driving systems - {P}art 2: Road users, animals, other obstacles,
  and environmental conditions},'' Waterloo Intelligent Systems Engineering Lab
  (WISE), Tech. Rep., 2018.

\bibitem{Meyer1992designByContract}
B.~Meyer, ``Applying {\textquotesingle}design by contract{\textquotesingle},''
  \emph{Computer}, vol.~25, no.~10, pp. 40--51, Oct. 1992.

\bibitem{Guo2017calibration}
C.~Guo, G.~Pleiss, Y.~Sun, and K.~Q. Weinberger, ``On calibration of modern
  neural networks,'' \emph{arXiv:1706.04599}, 2017.

\bibitem{Salay2018UsingMLSafely}
R.~Salay and K.~Czarnecki, ``Using machine learning safely in automotive
  software: An assessment and adaption of software process requirements in
  {ISO} 26262,'' \emph{arXiv:1808.01614}, 2018.

\bibitem{Schwaiger2020uncertainty}
A.~Schwaiger, P.~Sinhamahapatra, J.~Gansloser, and K.~Roscher, ``Is uncertainty
  quantification in deep learning sufficient for out-of-distribution
  detection?'' in \emph{Proc. of the Workshop on Artificial Intelligence Safety
  (AISafety)}, Yokohama, Japan, 2020.

\bibitem{Arnez2020_uq_comparison}
F.~Arnez, H.~Espinoza, A.~Radermacher, and F.~Terrier, ``A comparison of
  uncertainty estimation approaches in deep learning components for autonomous
  vehicle applications,'' \emph{arXiv:2006.15172}, 2020.

\bibitem{Piazzoni2020Modeling_conference}
A.~Piazzoni, J.~Cherian, M.~Slavik, and J.~Dauwels, ``Modeling sensing and
  perception errors towards robust decision making in autonomous vehicles,'' in
  \emph{Proc. of the 29th International Joint Conf. on Artificial Intelligence
  (IJCAI)}, 2020.

\bibitem{Weast2020}
J.~Weast, ``Sensors, safety models and a system-level approach to safe and
  scalable automated vehicles,'' \emph{arXiv:2009.03301}, 2020.

\bibitem{Klamann2019}
B.~Klamann, M.~Lippert, C.~Amersbach, and H.~Winner, ``Defining
  pass-/fail-criteria for particular tests of automated driving functions,'' in
  \emph{{IEEE} Intelligent Transportation Systems Conf. ({ITSC})}.\hskip 1em
  plus 0.5em minus 0.4em\relax {IEEE}, Oct. 2019.

\bibitem{Aravantinos2020}
V.~Aravantinos and P.~Schlicht, ``Making the relationship between uncertainty
  estimation and safety less uncertain,'' in \emph{Design, Automation {\&} Test
  in Europe Conf. {\&} Exhibition ({DATE})}.\hskip 1em plus 0.5em minus
  0.4em\relax {IEEE}, Mar. 2020.

\bibitem{Breitenstein2020visual}
J.~Breitenstein, J.-A. Term{\"o}hlen, D.~Lipinski, and T.~Fingscheidt,
  ``Systematization of corner cases for visual perception in automated
  driving,'' \emph{{IEEE} Intelligent Vehicles Symposium ({IV})}, 2020.

\bibitem{henne2020benchmarking}
M.~Henne, A.~Schwaiger, K.~Roscher, and G.~Weiss, ``Benchmarking uncertainty
  estimation methods for deep learning with safety-related metrics,'' in
  \emph{SafeAI@AAAI}, 2020, pp. 83--90.

\bibitem{Fleck2019testAreaBW}
T.~Fleck, K.~Daaboul, M.~Weber, P.~Sch{\"o}rner, M.~Wehmer, J.~Doll, S.~Orf,
  N.~Su{\ss}mann, C.~Hubschneider, M.~R. Zofka, F.~Kuhnt, R.~Kohlhaas,
  I.~Baumgart, R.~Z{\"o}llner, and J.~M. Z{\"o}llner, ``Towards large scale
  urban traffic reference data: Smart infrastructure in the test area
  autonomous driving {Baden-W{\"u}rttemberg},'' in \emph{Intelligent Autonomous
  Systems 15}, M.~Strand, R.~Dillmann, E.~Menegatti, and S.~Ghidoni, Eds.\hskip
  1em plus 0.5em minus 0.4em\relax Cham: Springer International Publishing,
  2019, pp. 964--982.

\bibitem{Spanfelner2012ChallengesISO26262}
B.~Spanfelner, D.~Richter, S.~Ebel, U.~Wilhelm, and C.~Patz, ``Challenges in
  applying the {ISO} 26262 for driver assistance systems,'' in \emph{Tagung
  Fahrerassistenz}, Munich, 2012.

\bibitem{Rahimi2019requirements}
M.~Rahimi, J.~L. Guo, S.~Kokaly, and M.~Chechik, ``Toward requirements
  specification for machine-learned components,'' in \emph{{IEEE} 27th
  International Requirements Engineering Conf. Workshops ({REW})}.\hskip 1em
  plus 0.5em minus 0.4em\relax {IEEE}, Sept. 2019.

\bibitem{Czarnecki2019devOps}
K.~Czarnecki, ``Software engineering for automated vehicles: Addressing the
  needs of cars that run on software and data,'' in \emph{{IEEE}/{ACM} 41st
  Int. Conf. on Software Engineering: Companion Proc.
  ({ICSE}-Companion)}.\hskip 1em plus 0.5em minus 0.4em\relax {IEEE}, May 2019.

\bibitem{Feng2018uncertainty}
D.~Feng, L.~Rosenbaum, and K.~Dietmayer, ``Towards safe autonomous driving:
  Capture uncertainty in the deep neural network for lidar {3D} vehicle
  detection,'' in \emph{21st Int. Conf. on Intelligent Transportation Systems
  ({ITSC})}.\hskip 1em plus 0.5em minus 0.4em\relax {IEEE}, Nov. 2018.

\bibitem{Feng2019calibrationECE}
D.~Feng, L.~Rosenbaum, C.~Glaeser, F.~Timm, and K.~Dietmayer, ``Can we trust
  you? {O}n calibration of a probabilistic object detector for autonomous
  driving,'' \emph{arXiv:1909.12358}, 2019.

\bibitem{NHTSA2017vision_for_safety2}
{USA DOT NHTSA}, ``Automated driving systems 2.0: A vision for safety,'' 2017.

\bibitem{Thorn2018NHTSA}
E.~Thorn, S.~C. Kimmel, and M.~Chaka, ``A framework for automated driving
  system testable cases and scenarios,'' USA DOT NHTSA, Tech. Rep., 2018.

\bibitem{UL4600_voting_2019}
\emph{UL4600 - Standard for Safety for the Evaluation of Autonomous Products},
  Underwriters Laboratories Std., Rev. Voting Version, Dec. 2019.

\bibitem{Koopman2019}
P.~Koopman, U.~Ferrell, F.~Fratrik, and M.~Wagner, ``A safety standard approach
  for fully autonomous vehicles,'' pp. 326--332, Aug. 2019.

\bibitem{UNECE2020lanekeepinglevel3}
\emph{Uniform provisions concerning the approval of vehicles with regard to
  Automated Lane Keeping Systems}, Economic Commission for Europe, Inland
  Transport Committee Std., Rev. ECE/TRANS/WP.29/2020/81, 2020.

\bibitem{SaFAD2019}
\BIBentryALTinterwordspacing
M.~Wood, P.~Robbel, M.~Maass, R.~Tebbens, M.~Meijs, M.~Harb, and P.~e.~a.
  Schlicht, ``Safety first for automated driving,'' Online, 2019. [Online].
  Available:
  \url{https://newsroom.intel.com/wp-content/uploads/sites/11/2019/07/Intel-Safety-First-for-Automated-Driving.pdf}
\BIBentrySTDinterwordspacing

\bibitem{ISO_TR_4804_2020}
``Iso/tr 4804 — road vehicles — safety and cybersecurity for automated
  driving systems — design, verification and validation,'' Dec. 2020.

\bibitem{Geiger2012CVPR}
\BIBentryALTinterwordspacing
A.~Geiger, P.~Lenz, and R.~Urtasun, ``Are we ready for autonomous driving?
  {T}he {KITTI} vision benchmark suite,'' in \emph{Conf. on Computer Vision and
  Pattern Recognition (CVPR)}, 2012. [Online]. Available:
  \url{http://www.cvlibs.net/datasets/kitti/eval\_tracking.php}
\BIBentrySTDinterwordspacing

\bibitem{caesar2019nuscenes}
H.~Caesar, V.~Bankiti, A.~H. Lang, S.~Vora, V.~E. Liong, Q.~Xu, A.~Krishnan,
  Y.~Pan, G.~Baldan, and O.~Beijbom, ``nu{S}cenes: A multimodal dataset for
  autonomous driving,'' \emph{arXiv:1903.11027}, 2019.

\bibitem{sun2019waymoopendataset}
P.~Sun, H.~Kretzschmar, X.~Dotiwalla, A.~Chouard, V.~Patnaik, P.~Tsui, J.~Guo,
  Y.~Zhou, Y.~Chai, B.~Caine, V.~Vasudevan, W.~Han, J.~Ngiam, H.~Zhao,
  A.~Timofeev, S.~Ettinger, M.~Krivokon, A.~Gao, A.~Joshi, Y.~Zhang, J.~Shlens,
  Z.~Chen, and D.~Anguelov, ``Scalability in perception for autonomous driving:
  Waymo open dataset,'' \emph{arXiv:1912.04838}, 2019.

\bibitem{Chang2019Argoverse3T}
M.-F. Chang, J.~Lambert, P.~Sangkloy, J.~Singh, S.~Bak, A.~T. Hartnett,
  D.~Wang, P.~Carr, S.~Lucey, D.~Ramanan, and J.~Hays, ``Argoverse: 3{D}
  tracking and forecasting with rich maps,'' \emph{IEEE/CVF Conf. on Computer
  Vision and Pattern Recognition (CVPR)}, pp. 8740--8749, 2019.

\bibitem{stampfle2005performance}
M.~Stampfle, D.~Holz, and J.~C. Becker, ``Performance evaluation of automotive
  sensor data fusion,'' in \emph{Proc. IEEE Intelligent Transportation
  Systems}.\hskip 1em plus 0.5em minus 0.4em\relax IEEE, 2005, pp. 50--55.

\bibitem{Kuhn1955_hungarianAlgo}
H.~W. Kuhn, ``The hungarian method for the assignment problem,'' \emph{Naval
  Research Logistics Quarterly}, vol.~2, no. 1-2, pp. 83--97, Mar. 1955.

\bibitem{Bertsekas1989_auctionAlgo}
D.~P. Bertsekas and D.~A. Castanon, ``The auction algorithm for the
  transportation problem,'' \emph{Annals of Operations Research}, vol.~20,
  no.~1, pp. 67--96, Dec. 1989.

\bibitem{bernardin2008evaluating}
K.~Bernardin and R.~Stiefelhagen, ``Evaluating multiple object tracking
  performance: the {CLEAR MOT} metrics,'' \emph{Journal on Image and Video
  Processing}, vol. 2008, p.~1, 2008.

\bibitem{Chen_2018}
\BIBentryALTinterwordspacing
Z.~Chen, C.~Heckman, S.~Julier, and N.~Ahmed, ``Weak in the {NEES}?:
  {A}uto-tuning kalman filters with bayesian optimization,'' \emph{21st Int.
  Conf. on Information Fusion (FUSION)}, July 2018. [Online]. Available:
  \url{http://dx.doi.org/10.23919/ICIF.2018.8454982}
\BIBentrySTDinterwordspacing

\bibitem{schuhmacher2008consistent}
D.~Schuhmacher, B.-T. Vo, and B.-N. Vo, ``A consistent metric for performance
  evaluation of multi-object filters,'' \emph{IEEE Transactions on Signal
  Processing}, vol.~56, no.~8, pp. 3447--3457, 2008.

\bibitem{ristic2010performance}
B.~Ristic, B.-N. Vo, and D.~Clark, ``Performance evaluation of multi-target
  tracking using the {OSPA} metric,'' in \emph{13th Int. Conf. on Information
  Fusion}.\hskip 1em plus 0.5em minus 0.4em\relax IEEE, 2010, pp. 1--7.

\bibitem{he2013track}
X.~He, R.~Tharmarasa, T.~Kirubarajan, and T.~Thayaparan, ``A track quality
  based metric for evaluating performance of multitarget filters,'' \emph{IEEE
  Transactions on Aerospace and Electronic Systems}, vol.~49, no.~1, pp.
  610--616, 2013.

\bibitem{vu2014new}
T.~Vu and R.~Evans, ``A new performance metric for multiple target tracking
  based on optimal subpattern assignment,'' in \emph{17th Int. Conf. on
  Information Fusion (FUSION)}.\hskip 1em plus 0.5em minus 0.4em\relax IEEE,
  2014, pp. 1--8.

\bibitem{beard2017ospa}
M.~Beard, B.~T. Vo, and B.-N. Vo, ``{OSPA} (2): Using the {OSPA} metric to
  evaluate multi-target tracking performance,'' in \emph{Int. Conf. on Control,
  Automation and Information Sciences (ICCAIS)}.\hskip 1em plus 0.5em minus
  0.4em\relax IEEE, 2017, pp. 86--91.

\bibitem{rahmathullah2017generalized}
A.~S. Rahmathullah, {\'A}.~F. Garc{\'\i}a-Fern{\'a}ndez, and L.~Svensson,
  ``Generalized optimal sub-pattern assignment metric,'' in \emph{20th Int.
  Conf. on Information Fusion (Fusion)}.\hskip 1em plus 0.5em minus 0.4em\relax
  IEEE, 2017, pp. 1--8.

\bibitem{shi2017comprehensive}
X.~Shi, F.~Yang, F.~Tong, and H.~Lian, ``A comprehensive performance metric for
  evaluation of multi-target tracking algorithms,'' in \emph{3rd Int. Conf. on
  Information Management (ICIM)}.\hskip 1em plus 0.5em minus 0.4em\relax IEEE,
  2017, pp. 373--377.

\bibitem{Reuter2014_LMB_paper}
S.~{Reuter}, B.~{Vo}, B.~{Vo}, and K.~{Dietmayer}, ``The labeled
  multi-bernoulli filter,'' \emph{IEEE Transactions on Signal Processing},
  vol.~62, no.~12, pp. 3246--3260, June 2014.

\bibitem{granstrom2016extended}
K.~Granstrom, M.~Baum, and S.~Reuter, ``Extended object tracking: Introduction,
  overview and applications,'' \emph{arXiv:1604.00970}, 2016.

\bibitem{li2009learning}
Y.~Li, C.~Huang, and R.~Nevatia, ``Learning to associate: Hybridboosted
  multi-target tracker for crowded scene,'' in \emph{IEEE Conf. on Computer
  Vision and Pattern Recognition}.\hskip 1em plus 0.5em minus 0.4em\relax IEEE,
  2009, pp. 2953--2960.

\bibitem{Luiten2020hota}
J.~Luiten, A.~Osep, P.~Dendorfer, P.~Torr, A.~Geiger, L.~Leal-Taixe, and
  B.~Leibe, ``{HOTA}: A higher order metric for evaluating multi-object
  tracking,'' \emph{International Journal of Computer Vision}, oct 2020.

\bibitem{Philion2020planner_centric}
J.~Philion, A.~Kar, and S.~Fidler, ``Learning to evaluate perception models
  using planner-centric metrics,'' in \emph{2020 {IEEE}/{CVF} Conference on
  Computer Vision and Pattern Recognition ({CVPR})}.\hskip 1em plus 0.5em minus
  0.4em\relax {IEEE}, June 2020.

\bibitem{Guo2020efficacy}
Y.~Guo, H.~Caesar, O.~Beijbom, J.~Philion, and S.~Fidler, ``The efficacy of
  neural planning metrics: A meta-analysis of pkl on nuscenes,''
  \emph{arXiv:2010.09350}, Oct. 2020.

\bibitem{Florbaeck2016.matching.offline}
J.~{Florbäck}, L.~{Tornberg}, and N.~{Mohammadiha}, ``Offline object matching
  and evaluation process for verification of autonomous driving,'' in
  \emph{IEEE 19th Int. Conf. on Intelligent Transportation Systems (ITSC)},
  Nov. 2016, pp. 107--112.

\bibitem{Sondell2018}
D.~Sondell and K.~Svensson, ``Matching vehicle sensors to reference sensors
  using machine learning methods,'' Master's thesis, Chalmers University of
  Technology, 2018.

\bibitem{hirsenkorn2015nonparametric}
N.~{Hirsenkorn}, T.~{Hanke}, A.~{Rauch}, B.~{Dehlink}, R.~{Rasshofer}, and
  E.~{Biebl}, ``A non-parametric approach for modeling sensor behavior,'' in
  \emph{16th International Radar Symposium (IRS)}, June 2015, pp. 131--136.

\bibitem{hanke2016classification}
T.~Hanke, N.~Hirsenkorn, B.~Dehlink, A.~Rauch, R.~Rasshofer, and E.~Biebl,
  ``Classification of sensor errors for the statistical simulation of
  environmental perception in automated driving systems,'' in \emph{IEEE 19th
  Int. Conf. on Intelligent Transportation Systems (ITSC)}.\hskip 1em plus
  0.5em minus 0.4em\relax IEEE, 2016, pp. 643--648.

\bibitem{Krajewski2020UsingDrones}
R.~Krajewski, M.~Hoss, A.~Meister, F.~Thomsen, J.~Bock, and L.~Eckstein,
  ``Using drones as reference sensors for neural-networks-based modeling of
  automotive perception errors,'' in \emph{IEEE Intelligent Vehicles Symposium
  (IV)}, 2020.

\bibitem{Zec2018markovModeling}
E.~L. Zec, N.~Mohammadiha, and A.~Schliep, ``Statistical sensor modelling for
  autonomous driving using autoregressive input-output {HMMs},'' in \emph{21st
  Int. Conf. on Intelligent Transportation Systems ({ITSC})}.\hskip 1em plus
  0.5em minus 0.4em\relax {IEEE}, Nov. 2018.

\bibitem{holder2019modeling}
M.~F. Holder, C.~Linnhoff, P.~Rosenberger, C.~Popp, and H.~Winner, ``Modeling
  and simulation of radar sensor artifacts for virtual testing of autonomous
  driving,'' in \emph{9. Tagung Automatisiertes Fahren}, 2019.

\bibitem{schaermann2017validation}
A.~Schaermann, A.~Rauch, N.~Hirsenkorn, T.~Hanke, R.~Rasshofer, and E.~Biebl,
  ``Validation of vehicle environment sensor models,'' in \emph{IEEE
  Intelligent Vehicles Symposium (IV)}.\hskip 1em plus 0.5em minus 0.4em\relax
  IEEE, 2017, pp. 405--411.

\bibitem{rosenberger2019towards}
P.~Rosenberger, J.~T. Wendler, M.~F. Holder, C.~Linnhoff, M.~Bergh{\"o}fer,
  H.~Winner, and M.~Maurer, ``Towards a generally accepted validation
  methodology for sensor models - challenges, metrics, and first results,''
  \emph{Proc. of Graz Symposium Virtual Vehicle (GSVF)}, 2019.

\bibitem{Holder2020radarModelPlanningInfluence}
M.~F. Holder, J.~R. Thielmann, P.~Rosenberger, C.~Linnhoff, and H.~Winner,
  ``How to evaluate synthetic radar data? {L}essons learned from finding
  driveable space in virtual environments,'' in \emph{13. Uni-DAS e.V. Workshop
  Fahrerassistenz und automatisiertes Fahren}, 2020.

\bibitem{Philipp2020DecompositionPerception}
R.~Philipp, F.~Schuldt, and F.~Howar, ``Functional decomposition of automated
  driving systems for the classification and evaluation of perceptual
  threats,'' in \emph{13. Uni-DAS e.V. Workshop Fahrerassistenz und
  automatisiertes Fahren}, 2020.

\bibitem{poddey2019opencontext}
A.~Poddey, T.~Brade, J.~E. Stellet, and W.~Branz, ``On the validation of
  complex systems operating in open contexts,'' \emph{arXiv:1902.10517}, 2019.

\bibitem{Gansch2020uncertainty}
R.~Gansch and A.~Adee, ``System theoretic view on uncertainties,'' in
  \emph{2020 Design, Automation {\&} Test in Europe Conference {\&} Exhibition
  ({DATE})}.\hskip 1em plus 0.5em minus 0.4em\relax {IEEE}, Mar. 2020.

\bibitem{Hu2020}
B.~C. Hu, R.~Salay, K.~Czarnecki, M.~Rahimi, G.~Selim, and M.~Chechik,
  ``Towards requirements specification for machine-learned perception based on
  human performance,'' in \emph{{IEEE} 7th International Workshop on Artificial
  Intelligence for Requirements Engineering ({AIRE})}.\hskip 1em plus 0.5em
  minus 0.4em\relax {IEEE}, Sept. 2020.

\bibitem{Dokhanchi2018qtl}
A.~Dokhanchi, H.~B. Amor, J.~V. Deshmukh, and G.~Fainekos, ``Evaluating
  perception systems for autonomous vehicles using quality temporal logic,'' in
  \emph{Runtime Verification}, C.~Colombo and M.~Leucker, Eds.\hskip 1em plus
  0.5em minus 0.4em\relax Cham: Springer International Publishing, 2018, pp.
  409--416.

\bibitem{Balakrishnan2019metrics}
A.~Balakrishnan, A.~G. Puranic, X.~Qin, A.~Dokhanchi, J.~V. Deshmukh, H.~B.
  Amor, and G.~Fainekos, ``Specifying and evaluating quality metrics for
  vision-based perception systems,'' in \emph{2019 Design, Automation {\&} Test
  in Europe Conference {\&} Exhibition ({DATE})}.\hskip 1em plus 0.5em minus
  0.4em\relax {IEEE}, Mar. 2019.

\bibitem{schoenemann2019fault}
V.~Schönemann, H.~Winner, G.~Verhaeg, F.~Tronci, and G.~A.~G. Padilla, ``Fault
  tree-based derivation of safety requirements for automated driving on the
  example of cooperative valet parking,'' 2019.

\bibitem{Pek2020onlineVerification}
C.~Pek, S.~Manzinger, M.~Koschi, and M.~Althoff, ``Using online verification to
  prevent autonomous vehicles from causing accidents,'' \emph{Nature Machine
  Intelligence}, vol.~2, no.~9, pp. 518--528, Sept. 2020.

\bibitem{Naeini2015calibrationError}
M.~P. Naeini, G.~F. Cooper, and M.~Hauskrecht, ``Obtaining well calibrated
  probabilities using bayesian binning,'' in \emph{Proc. of the AAAI Conf. on
  Artificial Intelligence}, vol. 2015.\hskip 1em plus 0.5em minus 0.4em\relax
  NIH Public Access, 2015, pp. 2901--2907.

\bibitem{Kuleshov2018regressionUncertainty}
V.~Kuleshov, N.~Fenner, and S.~Ermon, ``Accurate uncertainties for deep
  learning using calibrated regression,'' \emph{arXiv:1807.00263}, 2018.

\bibitem{Gustafsson2020calibrationError}
F.~K. Gustafsson, M.~Danelljan, and T.~B. Schon, ``Evaluating scalable bayesian
  deep learning methods for robust computer vision,'' in \emph{Proc. of the
  IEEE/CVF Conf. on Computer Vision and Pattern Recognition (CVPR) Workshops},
  June 2020.

\bibitem{Junietz2018criticalityMetric}
P.~Junietz, F.~Bonakdar, B.~Klamann, and H.~Winner, ``Criticality metric for
  the safety validation of automated driving using model predictive trajectory
  optimization,'' in \emph{21st Int. Conf. on Intelligent Transportation
  Systems ({ITSC})}.\hskip 1em plus 0.5em minus 0.4em\relax {IEEE}, Nov. 2018.

\bibitem{Avizienis2004dependability}
A.~Avizienis, J.-C. Laprie, B.~Randell, and C.~Landwehr, ``Basic concepts and
  taxonomy of dependable and secure computing,'' \emph{{IEEE} Transactions on
  Dependable and Secure Computing}, vol.~1, no.~1, pp. 11--33, Jan. 2004.

\bibitem{Rocklage2017scenarios}
E.~Rocklage, H.~Kraft, A.~Karatas, and J.~Seewig, ``Automated scenario
  generation for regression testing of autonomous vehicles,'' in \emph{{IEEE}
  20th Int. Conf. on Intelligent Transportation Systems ({ITSC})}.\hskip 1em
  plus 0.5em minus 0.4em\relax {IEEE}, Oct. 2017.

\bibitem{Menzel2018scenarios}
T.~Menzel, G.~Bagschik, and M.~Maurer, ``Scenarios for development, test and
  validation of automated vehicles,'' in \emph{{IEEE} Intelligent Vehicles
  Symposium ({IV})}.\hskip 1em plus 0.5em minus 0.4em\relax {IEEE}, June 2018.

\bibitem{Neurohr2021criticality}
C.~Neurohr, L.~Westhofen, M.~Butz, M.~Bollmann, U.~Eberle, and R.~Galbas,
  ``Criticality analysis for the verification and validation of automated
  vehicles,'' \emph{{IEEE} Access}, vol.~9, pp. 18\,016--18\,041, 2021.

\bibitem{bsi2020ODDStandard}
S.~Khastgir, \emph{Operational Design Domain (ODD) taxonomy for an automated
  driving system (ADS) – Specification}, The British Standards Institution
  Std. PAS 1883:2020, 2020.

\bibitem{Asam2020openODD}
\BIBentryALTinterwordspacing
{ASAM e. V.}, ``{ASAM OpenODD},'' 2020, accessed on 09/12/2020. [Online].
  Available: \url{https://www.asam.net/project-detail/asam-openodd/}
\BIBentrySTDinterwordspacing

\bibitem{Asam2020openXontology}
\BIBentryALTinterwordspacing
------, ``{ASAM OpenXOntology},'' 2020, accessed on 09/12/2020. [Online].
  Available: \url{https://www.asam.net/project-detail/asam-openxontology/}
\BIBentrySTDinterwordspacing

\bibitem{Schuldt2017diss}
\BIBentryALTinterwordspacing
F.~Schuldt, ``Ein {B}eitrag f{\"u}r den methodischen {T}est von automatisierten
  {F}ahrfunktionen mit {H}ilfe von virtuellen {U}mgebungen,'' Ph.D.
  dissertation, TU Braunschweig, Apr. 2017. [Online]. Available:
  \url{https://publikationsserver.tu-braunschweig.de/receive/dbbs_mods_00064747}
\BIBentrySTDinterwordspacing

\bibitem{Bock2018databasis}
J.~Bock, R.~Krajewski, L.~Eckstein, J.~Klimke, J.~Sauerbier, and A.~Zlocki,
  ``Data basis for scenario-based validation of had on highways,'' in
  \emph{27th Aachen Colloquium Automobile and Engine Technology}, 2018.

\bibitem{Scholtes2020_6LM}
M.~Scholtes, L.~Westhofen, L.~R. Turner, K.~Lotto, M.~Schuldes, H.~Weber,
  N.~Wagener, C.~Neurohr, M.~Bollmann, F.~Körtke, J.~Hiller, M.~Hoss, J.~Bock,
  and L.~Eckstein, ``6-layer model for a structured description and
  categorization of urban traffic and environment,'' \emph{arXiv:2012.06319},
  2020.

\bibitem{Driesten2019osi}
C.~van Driesten and T.~Schaller, ``Overall approach to standardize {AD} sensor
  interfaces: Simulation and real vehicle,'' in \emph{Fahrerassistenzsysteme
  2018}, T.~Bertram, Ed.\hskip 1em plus 0.5em minus 0.4em\relax Wiesbaden:
  Springer Fachmedien Wiesbaden, 2019, pp. 47--55.

\bibitem{Asam2020openSCENARIO}
\BIBentryALTinterwordspacing
{ASAM e. V.}, ``{ASAM OpenSCENARIO},'' 2021, accessed on 09/02/2021. [Online].
  Available: \url{https://www.asam.net/standards/detail/openscenario/}
\BIBentrySTDinterwordspacing

\bibitem{Menzel2019functionalToLogical}
T.~Menzel, G.~Bagschik, L.~Isensee, A.~Schomburg, and M.~Maurer, ``From
  functional to logical scenarios: Detailing a keyword-based scenario
  description for execution in a simulation environment,'' in \emph{{IEEE}
  Intelligent Vehicles Symposium ({IV})}.\hskip 1em plus 0.5em minus
  0.4em\relax {IEEE}, June 2019.

\bibitem{Neurohr2020fundamental}
C.~Neurohr, L.~Westhofen, T.~Henning, T.~de~Graaff, E.~Mohlmann, and E.~Bode,
  ``Fundamental considerations around scenario-based testing for automated
  driving,'' in \emph{{IEEE} Intelligent Vehicles Symposium ({IV})}.\hskip 1em
  plus 0.5em minus 0.4em\relax {IEEE}, Oct. 2020.

\bibitem{putz2017system}
A.~P{\"u}tz, A.~Zlocki, J.~Bock, and L.~Eckstein, ``System validation of highly
  automated vehicles with a database of relevant traffic scenarios,'' in
  \emph{12th ITS European Congr.}, vol.~1, 2017, pp. 19--22.

\bibitem{Krajewski2018traGAN}
R.~Krajewski, T.~Moers, D.~Nerger, and L.~Eckstein, ``Data-driven maneuver
  modeling using generative adversarial networks and variational autoencoders
  for safety validation of highly automated vehicles,'' in \emph{21st
  International Conference on Intelligent Transportation Systems
  ({ITSC})}.\hskip 1em plus 0.5em minus 0.4em\relax {IEEE}, Nov. 2018.

\bibitem{Schoener2020challenging}
{Hans-Peter Schoener}, ``\BIBforeignlanguage{en}{Challenging highway scenarios
  beyond collision avoidance for autonomous vehicle certification},''
  \emph{\BIBforeignlanguage{en}{ResearchGate preprint}}, Nov. 2020.

\bibitem{Kurzidem2020analyzingPerceptionArchitectures}
I.~Kurzidem, A.~Saad, and P.~Schleiss, ``A systematic approach to analyzing
  perception architectures in autonomous vehicles,'' in \emph{Model-Based
  Safety and Assessment}, M.~Zeller and K.~H{\"o}fig, Eds.\hskip 1em plus 0.5em
  minus 0.4em\relax Cham: Springer International Publishing, 2020, pp.
  149--162.

\bibitem{Kuhn2009combinatorialTesting}
R.~Kuhn, R.~Kacker, Y.~Lei, and J.~Hunter, ``Combinatorial software testing,''
  \emph{Computer}, vol.~42, no.~7, pp. 94--96, Aug. 2009.

\bibitem{Gladisch2020combinatorial}
C.~Gladisch, C.~Heinzemann, M.~Herrmann, and M.~Woehrle, ``Leveraging
  combinatorial testing for safety-critical computer vision datasets,'' in
  \emph{Proceedings of the IEEE/CVF Conference on Computer Vision and Pattern
  Recognition (CVPR) Workshops}, June 2020.

\bibitem{Weber2020simulation}
N.~Weber, D.~Frerichs, and U.~Eberle, ``A simulation-based, statistical
  approach for the derivation of concrete scenarios for the release of highly
  automated driving functions,'' in \emph{{AmE} - Automotive meets Electronics;
  11th GMM-Symposium}, Dortmund, Germany, 2020, pp. 1 -- 6.

\bibitem{Robosense2020reference}
\BIBentryALTinterwordspacing
Robosense, ``{RS-Reference} - efficiently access massive and reliable ground
  truth data,'' Online, Dec. 2020. [Online]. Available:
  \url{http://www.robosense.ai/en/rslidar/RS-Reference}
\BIBentrySTDinterwordspacing

\bibitem{Asam2020openLabel}
\BIBentryALTinterwordspacing
{ASAM e. V.}, ``{ASAM OpenLABEL},'' 2020, accessed on 09/12/2020. [Online].
  Available:
  \url{https://www.asam.net/project-detail/scenario-storage-and-labelling/}
\BIBentrySTDinterwordspacing

\bibitem{Scheiner2019gnssVRUs}
N.~Scheiner, N.~Appenrodt, J.~Dickmann, and B.~Sick, ``Automated ground truth
  estimation of vulnerable road users in automotive radar data using {GNSS},''
  in \emph{{IEEE} {MTT}-S Int. Conf. on Microwaves for Intelligent Mobility
  ({ICMIM})}.\hskip 1em plus 0.5em minus 0.4em\relax {IEEE}, Apr. 2019.

\bibitem{Ho2018rtkGpsAccuracy}
V.~Ho, K.~Rauf, I.~Passchier, F.~Rijks, and T.~Witsenboer, ``Accuracy
  assessment of {RTK} {GNSS} based positioning systems for automated driving,''
  in \emph{15th Workshop on Positioning, Navigation and Communications
  ({WPNC})}.\hskip 1em plus 0.5em minus 0.4em\relax {IEEE}, Oct. 2018.

\bibitem{Tahir2018_GPS_Accuracy}
M.~{Tahir}, S.~S. {Afzal}, M.~S. {Chughtai}, and K.~{Ali}, ``On the accuracy of
  inter-vehicular range measurements using {GNSS} observables in a cooperative
  framework,'' \emph{IEEE Transactions on Intelligent Transportation Systems},
  vol.~20, no.~2, pp. 682--691, Feb. 2019.

\bibitem{ESA2013gnss}
J.~S. Subirana, J.~J. Zornoza, and M.~Hern{\'a}ndez-Pajares, ``{GNSS} data
  processing. volume 1: Fundamentals and algorithms,'' \emph{ESA Communications
  ESTEC, PO Box}, vol. 299, p. 2200, 2013.

\bibitem{Lampe2019collectiveDriving}
B.~Lampe, T.~Woopen, and L.~Eckstein, ``{Collective Driving - Cloud Services
  for Automated Vehicles in UNICARagil},'' in \emph{28th Aachen
  {{Colloquium}}}, Aachen, Germany, Oct. 2019.

\bibitem{knake2016aimintersection}
S.~Knake-Langhorst, K.~Gimm, T.~Frankiewicz, and F.~Köster, ``Test site {AIM}
  – toolbox and enabler for applied research and development in traffic and
  mobility,'' \emph{Transportation Research Procedia}, vol.~14, pp. 2197--2206,
  Dec. 2016.

\bibitem{notz2020extraction}
D.~Notz, F.~Becker, T.~Kühbeck, and D.~Watzenig, ``Extraction and assessment
  of naturalistic human driving trajectories from infrastructure camera and
  radar sensors,'' \emph{arXiv:2004.01288}, 2020.

\bibitem{Kraemmer2019providentia}
A.~Krämmer, C.~Schöller, D.~Gulati, V.~Lakshminarasimhan, F.~Kurz,
  D.~Rosenbaum, C.~Lenz, and A.~Knoll, ``Providentia - a large-scale sensor
  system for the assistance of autonomous vehicles and its evaluation,''
  \emph{arXiv:1906.06789}, 2019.

\bibitem{Kloeker2020infrastructure_ACK}
L.~Kloeker, A.~Kloeker, F.~Thomsen, A.~Erraji, and L.~Eckstein, ``Traffic
  detection using modular infrastructure sensors as a data basis for highly
  automated and connected driving,'' \emph{29. Aachen Colloquium - Sustainable
  Mobility}, vol.~29, no.~2, pp. 1835--1844, 2020.

\bibitem{Kloeker2020trafficRecording}
L.~Kloeker, C.~Geller, A.~Kloeker, and L.~Eckstein, ``High-precision digital
  traffic recording with multi-lidar infrastructure sensor setups,''
  \emph{arXiv:2006.12140}, 2020.

\bibitem{Guido2016uav}
\BIBentryALTinterwordspacing
G.~Guido, V.~Gallelli, D.~Rogano, and A.~Vitale, ``Evaluating the accuracy of
  vehicle tracking data obtained from unmanned aerial vehicles,''
  \emph{International Journal of Transportation Science and Technology},
  vol.~5, no.~3, pp. 136 -- 151, 2016, {U}nmanned Aerial Vehicles and Remote
  Sensing. [Online]. Available:
  \url{http://www.sciencedirect.com/science/article/pii/S2046043016300272}
\BIBentrySTDinterwordspacing

\bibitem{Kruber2020uav}
F.~Kruber, E.~S. Morales, S.~Chakraborty, and M.~Botsch, ``Vehicle position
  estimation with aerial imagery from unmanned aerial vehicles,''
  \emph{arXiv:2004.08206}, 2020.

\bibitem{Krajewski2019dronesfortesting}
R.~Krajewski, J.~Bock, and L.~Eckstein, ``Drones as a tool for the development
  and safety validation of highly automated driving,'' in \emph{Aachener
  Kolloquium}, 2019, pp. 1449--1462.

\bibitem{kurz2018dlrad}
F.~Kurz, D.~Waigand, P.~Pekezou-Fouopi, E.~Vig, C.~Henry, N.~Merkle,
  D.~Rosenbaum, V.~Gstaiger, S.~Azimi, S.~Auer, \emph{et~al.}, ``{DLRAD} - a
  first look on the new vision and mapping benchmark dataset for autonomous
  driving,'' \emph{International Archives of the Photogrammetry, Remote Sensing
  \& Spatial Information Sciences}, vol.~42, no.~1, 2018.

\bibitem{Kurz2015helicopter}
F.~Kurz, D.~Rosenbaum, H.~Runge, and P.~Reinartz, ``Validation of advanced
  driver assistance systems by airborne optical imagery,'' \emph{Proc. of
  Mobil.TUM}, pp. 1--11, 2015.

\bibitem{wang2020inferring}
Z.~Wang, D.~Feng, Y.~Zhou, W.~Zhan, L.~Rosenbaum, F.~Timm, K.~Dietmayer, and
  M.~Tomizuka, ``Inferring spatial uncertainty in object detection,''
  \emph{arXiv:2003.03644}, 2020.

\bibitem{Wang2018formal}
S.~Wang, K.~Pei, J.~Whitehouse, J.~Yang, and S.~Jana, ``Efficient formal safety
  analysis of neural networks,'' \emph{arXiv:1809.08098}, Sept. 2018.

\bibitem{Seshia2016verifiedAI}
S.~A. Seshia, D.~Sadigh, and S.~S. Sastry, ``Towards verified artificial
  intelligence,'' \emph{arXiv:1606.08514}, 2020, v4 (after first version from
  2016).

\bibitem{Barbier2019PerceptionValidation}
M.~{Barbier}, A.~{Renzaglia}, J.~{Quilbeuf}, L.~{Rummelhard}, A.~{Paigwar},
  C.~{Laugier}, A.~{Legay}, J.~{Ibañez-Guzmán}, and O.~{Simonin},
  ``Validation of perception and decision-making systems for autonomous driving
  via statistical model checking,'' in \emph{IEEE Intelligent Vehicles
  Symposium (IV)}, 2019, pp. 252--259.

\bibitem{NeumannCosel2014vtd}
K.~von Neumann-Cosel, ``Virtual test drive: {S}imulation umfeldbasierter
  {F}ahrzeugfunktionen,'' Dissertation, Technical University Munich, Munich,
  2014.

\bibitem{Rao2019faultInjection}
D.~Rao, P.~Pathrose, F.~Huening, and J.~Sid, ``An approach for validating
  safety of perception software in autonomous driving systems,'' in
  \emph{Model-Based Safety and Assessment}, Y.~Papadopoulos, K.~Aslansefat,
  P.~Katsaros, and M.~Bozzano, Eds.\hskip 1em plus 0.5em minus 0.4em\relax
  Cham: Springer International Publishing, 2019, pp. 303--316.

\bibitem{Myklebust2020agile}
T.~Myklebust, T.~Stålhane, and G.~Hanssen, ``Agile safety case and
  \mbox{{DevOps}} for the automotive industry,'' in \emph{Proc. of the 30th
  European Safety and Reliability Conf. and the 15th Probabilistic Safety
  Assessment and Management Conf.}, Aug. 2020.

\bibitem{Antonante2020perceptionMonitoringv3}
P.~Antonante, D.~I. Spivak, and L.~Carlone, ``Monitoring and diagnosability of
  perception systems,'' \emph{arXiv:2011.07010}, 2020, v3.

\bibitem{Woopen2018UNICARagil}
T.~Woopen, B.~Lampe, T.~B\"oddeker, L.~Eckstein, A.~Kampmann, B.~Alrifaee,
  S.~Kowalewski, D.~Moormann, T.~Stolte, I.~Jatzkowski, M.~Maurer, M.~M\"ostl,
  R.~Ernst, S.~Ackermann, C.~Amersbach, S.~Leinen, H.~Winner, D.~P\"ullen,
  S.~Katzenbeisser, M.~Becker, C.~Stiller, K.~Furmans, K.~Bengler,
  F.~Diermeyer, M.~Lienkamp, D.~Keilhoff, H.-C. Reuss, M.~Buchholz,
  K.~Dietmayer, H.~Lategahn, N.~Siepenk\"otter, M.~Elbs, E.~v.~Hin\"uber,
  M.~Dupuis, and C.~Hecker, ``{UNICARagil - Disruptive Modular Architectures
  for Agile, Automated Vehicle Concepts},'' in \emph{27th Aachen
  {{Colloquium}}}, Aachen, Germany, Oct. 2018.

\bibitem{Nolte2020selfRepresentation}
M.~Nolte, I.~Jatzkowski, S.~Ernst, and M.~Maurer, ``Supporting safe decision
  making through holistic system-level representations \& monitoring -- a
  summary and taxonomy of self-representation concepts for automated
  vehicles,'' \emph{arXiv:2007.13807}, 2020.

\bibitem{Lampe2020reducing}
B.~Lampe, R.~van Kempen, T.~Woopen, A.~Kampmann, B.~Alrifaee, and L.~Eckstein,
  ``Reducing uncertainty by fusing dynamic occupancy grid maps in a cloud-based
  collective environment model,'' in \emph{{IEEE} Intelligent Vehicles
  Symposium ({IV})}.\hskip 1em plus 0.5em minus 0.4em\relax {IEEE}, Oct. 2020.

\bibitem{Winner2016quoVadis}
\BIBentryALTinterwordspacing
H.~Winner, \emph{ADAS, Quo Vadis?}\hskip 1em plus 0.5em minus 0.4em\relax Cham:
  Springer International Publishing, 2016, pp. 1557--1584. [Online]. Available:
  \url{https://doi.org/10.1007/978-3-319-12352-3_62}
\BIBentrySTDinterwordspacing

\bibitem{Grubmuller2019robustFusion}
S.~Grubmuller, G.~Stettinger, M.~A. Sotelo, and D.~Watzenig, ``Fault-tolerant
  environmental perception architecture for robust automated driving,'' in
  \emph{{IEEE} Int. Conf. on Connected Vehicles and Expo ({ICCVE})}.\hskip 1em
  plus 0.5em minus 0.4em\relax {IEEE}, Nov. 2019.

\end{thebibliography}

\end{document}